\documentclass[10pt,twocolumn,letterpaper]{article}

\usepackage{cvpr}
\usepackage{times}
\usepackage{epsfig}
\usepackage{graphicx}
\usepackage{amsmath}
\usepackage{amssymb}

\usepackage{paralist}
\usepackage{multirow}
\usepackage[font=small]{caption}
\usepackage[square,numbers,sort&compress]{natbib}
\usepackage{xcolor}

\usepackage{paralist}

\newcommand{\abc}[1]{\textcolor{black}{#1}}
\newcommand{\abcn}[1]{\textcolor{black}{#1}}
\newcommand{\abcnn}[1]{\textcolor{black}{#1}}
\newcommand{\CUT}[1]{}

\newcommand{\calC}{{\mathcal C}}
\newcommand{\calD}{{\mathcal D}}
\newcommand{\diag}{\text{diag}}

\usepackage{multirow}

\newcommand{\wqz}[1]{\textcolor{black}{#1}}
\definecolor{yellowgreen}{rgb}{0.6, 0.8, 0.2}
\newcommand{\wqzn}[1]{\textcolor{black}{#1}}
\newcommand{\wqznn}[1]{\textcolor{black}{#1}}

\usepackage[pagebackref=true,breaklinks=true,letterpaper=true,colorlinks,bookmarks=false]{hyperref}

\cvprfinalcopy 

\def\cvprPaperID{1964} 

\ifcvprfinal\fi
\begin{document}

\title{
Describing like Humans: on Diversity in Image Captioning
}

\author{
Qingzhong Wang and Antoni B. Chan\\
Department of Computer Science \\City University of Hong Kong\\
{\tt\small qingzwang2-c@my.cityu.edu.hk, \tt\small abchan@cityu.edu.hk}
}

\maketitle

\begin{abstract}
Recently, the state-of-the-art models for image captioning have overtaken human performance based on the most popular metrics, such as BLEU, METEOR, ROUGE and CIDEr. Does this mean we have solved the task of image captioning? 
The above metrics only measure the similarity of the generated caption to the human annotations, which reflects its accuracy.
\abcn{However, an image contains many concepts and multiple levels of detail, and thus there is a variety of captions that express different concepts and details that might be interesting for different humans.}
%
%
Therefore only evaluating accuracy is not sufficient for measuring the performance of captioning models -- the diversity of the generated captions should also be considered.
%
\abcn{In this paper, we proposed a new metric for measuring diversity of image captions, 
which is derived from latent semantic analysis and kernelized to use CIDEr similarity.}
%
 %
\abcn{We conduct extensive experiments to re-evaluate recent captioning models in the context of both diversity and accuracy.  We find that there is still a large gap between the model and human performance in terms of both accuracy and diversity, and that models that have optimized accuracy (CIDEr) have low diversity.}
\abcn{We also show that balancing the cross-entropy loss and CIDEr reward in reinforcement learning during training can effectively control the tradeoff between diversity and accuracy of the generated captions.}
\end{abstract}

\begin{figure}[t]
\centering
\includegraphics[width=0.5\textwidth]{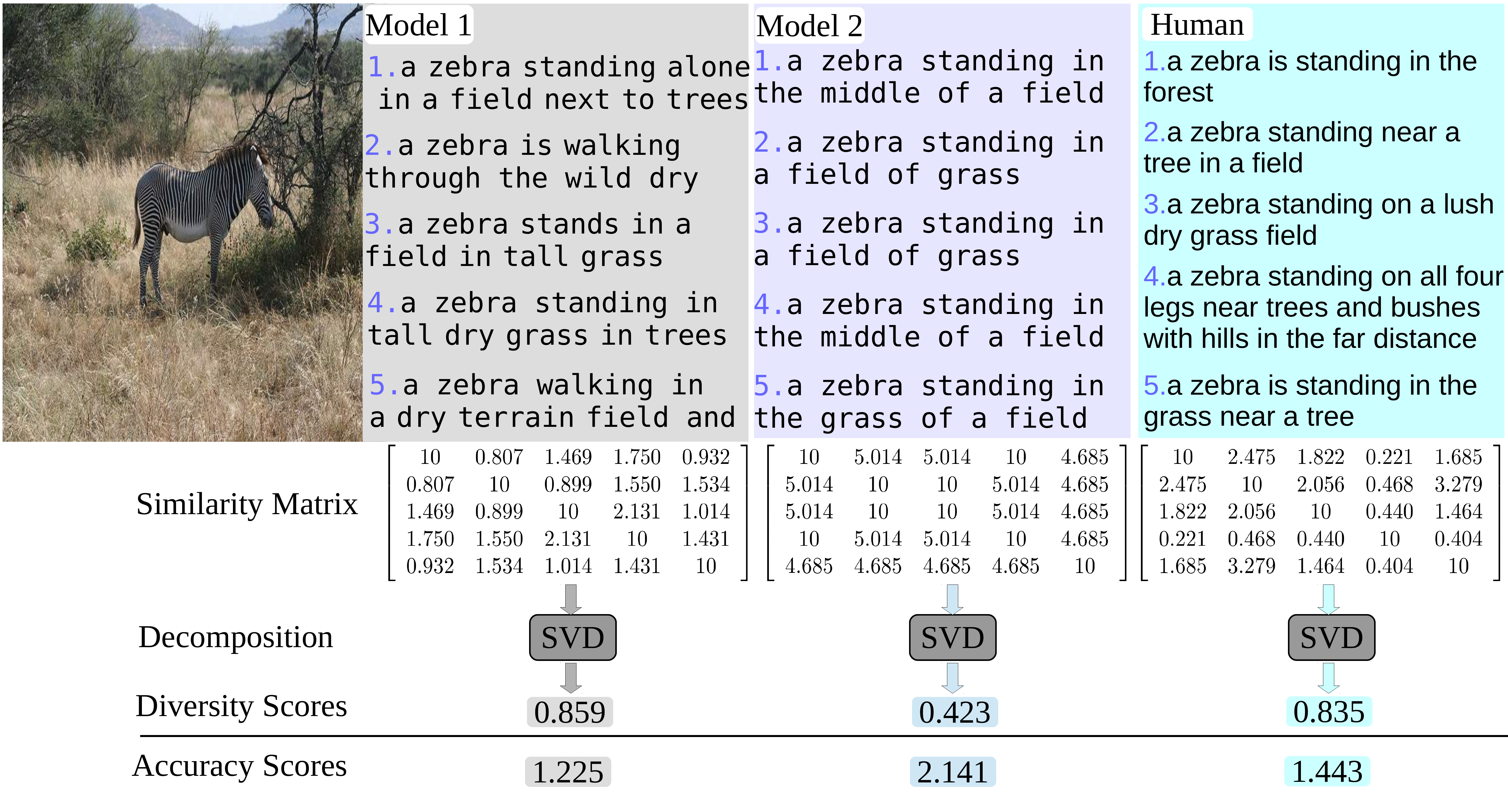}
\vspace{-2em}
\caption{\wqz{An overview of our diversity  metric. Given a set of captions \abcnn{from a method}, 
we first construct the \abcnn{self-}similarity matrix $\mathbf{K}$, 
\abcn{consisting of CIDEr \cite{C} scores between all pairs of captions}.
The diversity score is computed from the singular values of $\mathbf{K}$.
\abcn{A higher diversity score indicates more variety in the set of generated captions, such as changes in the level of descriptive detail and inclusion of removal of objects.}
\abcnn{The accuracy (average CIDEr) of the captions with respect to the human ground-truth is on the bottom.  For human annotations, this is the leave-one-out accuracy.}
}}\label{fig1}
\vspace{-1.1cm}
\end{figure}

\section{Introduction} \label{intro}
The task of image captioning is challenging and draws much attention from researchers in the fields of both computer vision and natural language processing.  A large variety of models have been proposed to automatically generate image captions, and most of the models are engaged in improving the accuracy of the generated captions as measured by the current metrics, such as BLEU 1-4 \cite{bleu}, METEOR \cite{M}, ROUGE \cite{R}, CIDEr \cite{C} and SPICE \cite{spice}.  However, another important property, the \textbf{diversity} of captions \abc{generated for a given image}, receives less attention. Generally, \textbf{diversity} refers to the differences among a set of captions generated by a method for a single image, and can be categorized into three levels:
(1) {\em word diversity} refers to only changes of single words that do not change the caption's semantics, \eg, using synonyms in different captions;  
(2) {\em syntactic diversity} refers to only differences in the word order, phrases, and sentence structures, such as pre-modification, post-modification, redundant and concise descriptions, which do not change the caption's concept. 
(3) {\em semantic diversity} refers to the differences of expressed concepts, \abc{including level of descriptive detail, changing of the sentence's subject, and addition/removal of sentence objects.}
\abc{For example, in Figure \ref{fig1}, the human captions 2 and 5 have syntactic diversity, as they both express the same concept of a zebra near a tree in a grass field using different word orderings.  In contrast, captions 2 and 3 exhibit semantic diversity, as caption 3 describes the type of grass field (``lush dry") and omits ``near a tree" as unimportant.
Ideally, a caption system should be able to generate captions expressing different concepts in the image, and hence in this paper we focus on measuring {\em semantic diversity}.}

The motivations for considering diversity of image captions are as follows.
First, an image may contain many concepts with multiple levels of detail --- indeed, 
{\em an image is worth a thousand words} --- and thus an image caption  expresses a set of concepts that are interesting for a particular human.   
Hence, there is diversity among captions due to diversity among humans, and an automatic image captioning method should reflect this. 
Second, only focusing on increasing the caption accuracy will bias the captioning method to common phrases.
For example, Figure \ref{fig1} shows the set of captions generated by two models. 
Model 2 is the best when only considering accuracy. 
However, Model 2 just repeats the same common phrases, \abc{providing no particular additional details}. In contrast, Model 1 recognizes that there are trees in the image and the the grass is dry, which also occurs in the human annotations. \wqz{It even recognizes ``walking'', which does not appear in the human annotations, but is a plausible description.}
Thus, to mimic the ability of humans, the captioning models should also have the ability of generating diverse captions.
Third, from the machine learning viewpoint, captioning models are typically trained on datasets where each image has at least 5 ground-truth captions  (e.g., MSCOCO), and thus captioning models should also be evaluated on how well the learned conditional distributions of captions given image approximates that of the ground-truth.
\abcnn{In particular, while the caption accuracy measures the differences in the modes of the distributions, the caption diversity measures the variance of the distribution.}

%

Recently, while a few works have focused on generating both diverse and accurate captions, such as conditional variational auto-encoders (CVAE) \cite{cvae} and conditional generative adversarial network (CGAN) \cite{cgan,cgan1}, there is not a metric to well evaluate the diversity of captions. 
%
%
%
In \cite{cgan}, the diversity of captions is shown only qualitatively. \cite{cvae,cgan1} evaluate the diversity of captions in three ways: 1) the percentage of novel sentences; 2) the percentage of unique uni-grams and bi-grams in the set of captions; 
3) 
\wqz{mBLEU,
\abcn{which is the average of the BLEU scores between each caption and the remaining captions.}}
However, it is difficult to define a novel sentence, and only considering the percentage of \abcn{unique} uni-grams and bi-grams ignores the \wqz{relationship} between captions, 
\abcn{e.g., the same n-gram could be used to construct sentences with different meanings.}
\abcnn{Because mBLEU uses the BLEU score, it aggregates n-grams over all the remaining captions, which obfuscates differences among the individual captions, thus under-representing the diversity.}
\wqzn{For example, the two caption sets, $\calC_1=$\{\textit{``zebras grazing grass''}, \textit{``grazing grass''}, \textit{``zebras grazing''}\} and $\calC_2=$\{\textit{``zebras grazing''}, \textit{``zebras grazing''}, \textit{``zebras grazing''}\}, obtain the same mBLEU of 1.0. However, we may consider that $\calC_1$ is more diverse, because each of its captions expresses different concepts or details.  In contrast, captions in $\calC_2$ describe exactly the same thing.} 
%
Hence, considering all the pairwise relationships among  captions will better 
reflect the structure of the set of captions. 
\abc{Moreover, BLEU is not a good metric for measuring semantic differences, since phrase-level changes and semantic changes may lead to the same BLEU score (e.g., see Table \ref{tab1})}.


In this paper, we propose a diversity measure based on pairwise similarities between captions. \abcn{In particular, we form a matrix of pairwise similarities (e.g., using CIDEr), and then use the singular values of the matrix to measure the diversity.  We show that this is interpretable as applying latent semantic analysis (LSA) on the weighted n-gram feature representation of the captions to extract the topic-structure of the set of captions, where more topics indicates more diversity in the captions.}
The key contributions of this paper are three-fold: 1) we proposed a new metric for evaluating diversity of sets of captions, and we re-evaluate existing captioning models via considering both diversity and accuracy; 
(2) we develop a framework that enables a tradeoff between diverse and accurate captions via balancing the rewards in reinforcement learning (RL) and the cross-entropy loss;
(3) extensive experiments are conducted to demonstrate the effectiveness of the diversity metric and the effect of the loss function on diversity and accuracy -- we find that RL and adversarial training are different approaches that provide equally satisfying results.

\section{Related Work} \label{rw}

\begin{table*}[t]
\centering
\small
\scalebox{0.95}{
\begin{tabular}{l|p{2.0in}|cccc|c|c|c|c}
\hline
Modification & Caption &B1 &B2 &B3 &B4 &M &R &C/10 &S \\
\hline
Reference &a group of people are playing football on a grass covered field &1 &1 &1 &1 &1 &1 &1 &1\\
\hline\hline
Word-level &a \textbf{couple} of \textbf{boys} are playing \textbf{soccer} on a grass covered field &0.750 &0.584 &0.468 &0.388 &0.387 &0.750 &0.261 &0.333\\
\hline
Phrase-level &\textbf{some guys} are playing football on a \textbf{grassy ground} &0.417 &0.389 &0.357 &0.317 &0.310 &0.489 &0.441 &0.133\\
\hline
Sentence-level &\textbf{on a grass covered field} a group of people are playing football &1.000 &0.953 &0.899 &0.834 &0.581 &0.583 & 0.676 &0.941\\
\hline
Redundancy &a group of people \textbf{in red soccer suits} are playing football on a grass covered field &0.716 &0.683 &0.644 &0.598 &0.429 &0.836 & 0.496 &0.818\\
\hline
Conciseness &a group of people are playing football &0.583 &0.564 &0.542 &0.516 &0.526 &0.774 & 0.482 &0.714\\
\hline\hline
Average &  &0.693 &0.635 &0.582 &0.531 &0.447 &0.693 &0.471 &0.588\\
\hline\hline
Semantic change &a group of people are \textbf{watching TV} &0.417 &0.389 &0.357 &0.317 &0.270 &0.553 &0.072 &0.429\\
\hline
\end{tabular}
}
\caption{The similarity scores between a reference caption and a modified caption using different evaluation metrics.
The caption in the first row is the reference caption, and the next five captions change different parts of the sentence (highlighted in bold) while keeping the same concepts.
%
\abc{``Average'' is the average metric value over these 5 modified captions.}
The bottom row shows an incorrect caption and the metric scores.
\abc{B1-4, M, R, C/10, and S are BLEU1-4, METEOR, ROUGE, CIDEr divided by 10 (so that the maximum is 1), and SPICE.
}
}\label{tab1}
\vspace{-1.5em}
\end{table*}

\textbf{Image Captioning.} Early image captioning models normally contain 2 stages: 1) concept detection, 2) translation. In the first stage, object categories, attributes and activities are detected, then the translation stage uses the labels to generate sentences. The typical concept detection models are conditional random fields (CRFs) \cite{templatemodel,babytalk}, support vector machines (SVMs) \cite{ngrammodel} or convolutional neural networks (CNNs) \cite{visconcept}, and the translation model is a sentence template \cite{templatemodel} or n-gram model \cite{ngrammodel}.

Recently, the encoder-decoder models, \wqz{e.g., neural image captioning (NIC) \cite{NIC}, spatial attention \cite{spatt} and adaptive attention \cite{when2look},} trained end-to-end 
have obtained much better results than the early models \wqz{based on concept detection and translation}. 
NIC \cite{NIC} translates images into sentences via directly connecting the inception network to an LSTM. To improve NIC, \cite{spatt} introduces a spatial attention module, which allows the model to ``watch'' different areas when it predicts different words. \cite{semanticatt,attributes,scn,attributesboosting} use image semantics detected using an additional network branch.
In \cite{when2look}, a sentinel gate decides whether the visual feature or the semantic feature should be used for prediction. 
Instead of employing LSTM decoders for sentences, \cite{convimagecap, cnnpluscnn, gha} apply convolutional decoders, which achieves faster training process and comparable results.

Both LSTM  and convolutional models are trained using cross-entropy. 
In contrast,  \cite{scst,bmcr} directly improve the evaluation metric using reinforcement learning (RL). They also show that improving the CIDEr reward function also improves other evaluation metrics, but not vice versa. 
Instead of using metric rewards, \cite{disccap,disccap2} employ the retrieval reward to generate more distinctive captions.

Generally, the above models are used to generate a single caption for one image, whereas \cite{cgan,cgan1} use 
CGAN to generate a set of diverse captions for each image.  The generator uses an LSTM to generate captions given an image, and the evaluator uses a retrieval model to evaluate the generated captions.  
The generator and evaluator are jointly trained in adversarial manner using policy gradients\footnote{This is similar to RL models, but RL models are trained by maximizing the rewards, which is not adversarial training.}. In the inference stage,  latent noise vectors are sampled from a 
Gaussian distribution, generating different captions.
 CVAE \cite{cvae} is another model that is able generate diverse caption by sampling the latent noise vector. 

\textbf{Evaluation Metrics.}
The most popular metrics are BLEU \cite{bleu}, METEOR \cite{M}, ROUGE \cite{R}, which are metrics from machine translation and document summarization, and 
CIDEr \cite{C} and SPICE \cite{spice}, 
which are metrics specific to image captioning.
BLEU, METEOR, ROUGE and CIDEr are based on computing the overlap between the n-grams of a generated caption and those of the 
human annotations.
%
BLEU considers the n-gram precision, ROUGE is related to n-gram recall, which benefits long texts, and METEOR takes both precision and recall of unigrams, while also applying synonym matching.
CIDEr uses TF-IDF weighted n-grams to represent captions and calculates the cosine similarity. 

Only considering n-gram overlap seems to ignore semantics of the captions.
SPICE uses scene graphs \cite{scenegraph,text2scenegraph} to represent images -- human annotations and one generated caption are first parsed into scene graphs, which are composed of object categories, attributes and relationships, and the F1-measure is computed between \wqz{the two scene graphs}.
However, SPICE is highly dependent on the accuracy of the parsing results. 
\cite{evalWMD} proposed a metric based on word2vec \cite{word2vec} and word mover distance (WMD) \cite{WMD}, which could leverage semantic information, but depends on the quality of word2vec. Recently, \cite{learn2eval} proposed a learned metric that uses a CNN and an LSTM to extract features from images and captions, and then uses a classifier to assign a score that indicates whether the caption is generated by a human.
While this metric is robust, it requires training and data augmentation, and the evaluation procedure takes more time.

Table \ref{tab1} shows an example of similarity metrics between a reference caption and 5 modified captions that have the same semantic meaning, and an incorrect caption with different meaning.
All metrics are less sensitive (have higher values) to sentence-level changes (due to the use of n-grams), in particular BLEU, ROUGE and SPICE.
Furthermore, all metrics show sensitivity to word-level or phrase-level changes.
%
Overall, CIDEr and METEOR have relatively low average metric value, which means that they are sensitive to sentence changes that keep the same semantics.
On the other hand, CIDEr and METEOR also assign lower values to the incorrect caption that changes the semantic meaning, which indicates that they are better able to discriminate between semantic changes in the sentence.
Hence, in this paper, we mainly consider CIDEr as the baseline metric to evaluate both the diversity and accuracy.


\section{Measuring Diversity of Image Captions} \label{de}

Currently, the widely used metrics, such as BLEU, CIDEr, and SPICE are for a single caption prediction.
To evaluate a set of captions $\calC=\{c_1, c_2, \cdots, c_m\}$, two dimensions are required: accuracy and diversity. For accuracy, \abc{the standard approach} is to average the similarity scores, $acc = \frac{1}{m} \sum_i s_i$, 
where $s_i=sim(c_i, \calC_{GT})$ is the similarity measure (e.g., CIDEr) between caption $c_i$ and ground-truth caption set $\calC_{GT}$.
%
For diversity, we will consider the pairwise similarity between captions in $\calC$, which is able to reflect the underlying structure of the set of captions.

\subsection{Latent Semantic Analysis} \label{LSA}
Latent semantic analysis (LSA) \cite{lsa} is a linear representation model, which is widely applied in information retrieval.  LSA considers the co-occurrence information of words (or n-grams), and uses singular value decomposition (SVD) to obtain a low-dimensional representations of the documents in terms of topic vectors. \abc{Applying LSA to a caption set}, more topics indicates a more diverse set of caption, whereas only one topic indicates a non-diverse set.
To use LSA, we  first represent each caption via a vector.  In this subsection, we consider the simplest representation, bag-of-words (BoW), and kernelize it in the next subsection using CIDEr.

 Given a set of captions $\calC=\{c_1, \cdots, c_m\}$ that describe an image, and a dictionary $\calD=\{w_1, w_2, \cdots, w_d\}$, we use the word-frequency vector to represent each caption $c_i$, 
 $\mathbf{f}_i = [f_1^i, \cdots, f_d^i]^T$, where $f_j^i$ denotes the frequency of word $w_j$ occurring in caption  $c_i$.
 %
 %
The caption set $\calC$ can be represented by a ``word-caption'' matrix, 
$\mathbf{M} = [\mathbf{f}_1 \cdots \mathbf{f}_m]$.


Applying SVD, we decompose $\mathbf{M}$ into three matrices, \ie, $\mathbf{M}=\mathbf{U}\mathbf{S}\mathbf{V}^T$, where $\mathbf{U}$ is composed of the eigenvectors of $\mathbf{M}\mathbf{M}^T$ and $\mathbf{S}=\diag(\sigma_1,\cdots, \sigma_m)$ is a diagonal matrix consisting of singular values $\sigma_1>\sigma_2>\cdots>0$ , and $\mathbf{V}$ is composed of the eigenvectors of $\mathbf{M}^T\mathbf{M}$. 
%
Each column of $\mathbf{U}$ represents \abc{the words in a topic vector of the caption set}, 
while the singular values in $\mathbf{S}$ represent the strength (frequency) of the topics.
If all captions in $\calC$ are the same, then only one singular value is non-zero, i.e., $\sigma_1>0$ and $\sigma_i=0, \forall i>1$,
If all the captions are different, then 
all the singular values are the same, i.e., $\sigma_1=\sigma_i, \forall i$. Hence, the ratio $r=\frac{\sigma_1}{\sum_{i=1}^{m}\sigma_i}$ 
%
%
represents how diverse the captions are, with larger $r$ meaning less diverse (i.e., the same caption), and smaller $r$ indicating more diversity (all different captions).
The ratio $r$ is within $[\frac{1}{m}, 1]$. 
Thus we map the ratio to a value in $[0,1]$, to obtain our diversity score $div=-\log_m(r)$, where larger $div$ means higher diversity.


Looking at the matrix $\mathbf{K}=\mathbf{M}^T\mathbf{M}$, each element $k_{ij} = \mathbf{f}_i^T\mathbf{f}_j$ 
is the dot-product similarity between the BoW vectors $\mathbf{f}_i$ and $\mathbf{f}_j$.
\abc{As the dimension of $\mathbf{f}_i$ may be large, a more efficient approach to computing the singular values is to use}
the eigenvalue decomposition 
$\mathbf{K}=\mathbf{V}\mathbf{\Lambda}\mathbf{V}^T$, where $\mathbf{\Lambda}=\diag(\lambda_1, \cdots, \lambda_m)$ are the eigenvalues of $\mathbf{K}$, which are the squares of the singular values, $\sigma_i = \sqrt{\lambda_i}$.
Note that $\mathbf{K}$ is a kernel matrix, and here LSA is using the linear kernel.

\subsection{Kernelized Method via CIDEr}\label{CIDEr}

In Section \ref{LSA}, a caption is represented by BoW features $\mathbf{f}_i$. 
However, this only considers word frequency and ignores phrases and sentence structures.
To address this problem, we use n-gram or p-spectrum kernels \cite{kernelbook} with LSA. The mapping function from the caption space $\mathbb{C}$ to the feature space $\mathbb{F}$ associated with the n-gram kernel is 
\begin{align}
\phi^n(c) = [f_1^n(c) \cdots  f_{|\calD^n|}^n(c)]^T, 
\end{align}
where $f_i^n(c)$ is the frequency of the $i$-th $n$-gram in caption $c$, and $\calD^n$ is the $n$-gram dictionary. 

CIDEr first projects the caption $c\in \mathbb{C}$ into a weighted feature space $\mathbb{F}$, $\Phi^n(c)=[\omega_i^n f_i^n(c)]_i$ 
where the weight $\omega_i^n$ for the $i$-th $n$-gram 
is its inverse document frequency (IDF). The CIDEr score is the average of the cosine similarities for each $n$,
\begin{align}\label{eq4}
CIDEr(c_i,c_j) = \frac{1}{4}\sum_{n=1}^{4}CIDEr_n(c_i, c_j),
\end{align}
where 
\begin{align}\label{eq5}
CIDEr_n(c_i,c_j) = \frac{\Phi^n(c_i)^T \Phi^n(c_j)}{||\Phi^n(c_i)||\ ||\Phi^n(c_j)||}.
\end{align}
%
In (\ref{eq5}),  $CIDEr_n$ is written as the cosine similarity kernel and the corresponding feature space is spanned by $\Phi^n(c)$. 
Since $CIDEr$ is the average of $CIDEr_{n}$ for different $n$, therefore, it is also a kernel function that accounts for uni-, bi-, tri- and quad-grams.

Since CIDEr can be interpreted as a kernel function, 
we reconsider the kernel matrix $\mathbf{K}$ in LSA, by using $k_{ij}=CIDEr(c_i,c_j)$.
The diversity according to CIDEr can then be computed 
by finding the eigenvalues of the kernel matrix $\{\lambda_1,\cdots,\lambda_m\}$, computing the  ratio $r=\frac{\sqrt{\lambda_1}}{\sum_{i=1}^m\sqrt{\lambda_i}}$, and applying the mapping function, $div=-\log_m(r)$.
\abc{Here, we are computing the diversity by using LSA to find the caption topics in the weighted n-gram feature space, rather than the original BoW space.  Other caption similarity measures could also be used in our framework to compute diversity if they can be written as \abcnn{positive definite} kernel functions.}

%

\section{Experiment Setup}

\abcn{We next present our experiment setup re-evaluating current captioning methods using both diversity and accuracy.
}

\subsection{Generating Diverse Captions} 
As most current models are trained to generate a single caption, 
we first must adapt them to generate a set of diverse captions.
In this paper we propose 4 approaches to generate diverse captions from a baseline model.
 (1) \textbf{Random sampling (RS):} 
After training, a set of captions is generated by randomly sampling word-by-word from the learned conditional distribution $\hat{p}(c|I)$.
(2) \textbf{Randomly cropped images (RCI):} 
The image is resized to $256\times 256$, and then randomly cropped to $224\times 224$ as input to generate the caption. 
(3) \textbf{Gaussian noise corruption (GNC):}
Gaussian noise with different standard deviations is added to the input image when predicting the caption.
(4) \textbf{Synonym switch (SS):}
The above 3 approached manipulate images to generate diverse captions, whereas the synonym switch approach directly manipulates a generated caption. First, a word2vec \cite{word2vec} model is trained on MSCOCO.
\abcn{
Next, given a caption, the top-10 synonyms for each word are retrieved and given a weight based on the similarities of their word2vec representation. Finally, with probability $p$, each word is randomly switched with one of its 10 synonyms, where the synonyms are sampled according to their weights.}

For the models that are able to generate diverse captions, such as CVAE and CGAN, \textbf{different random vectors (DRV)} are drawn from Gaussian distributions with different standard deviations to generate the captions. 

\subsection{Implementation Details}
In this paper, we evaluate the following captioning models: (1) NIC \cite{NIC} with VGG16 \cite{vggnet}; (2) SoftAtt \cite{spatt} with VGG16; (3) AdapAtt \cite{when2look} with VGG16; (4) Att2in \cite{scst} with cross-entropy (XE) and CIDEr reward, denoted as Att2in(XE) and Att2in(C); (5) FC \cite{scst} with cross-entropy and CIDEr reward, denoted as FC(XE) and FC(C); (6) Att2in and FC with retrieval reward\footnote{
\url{https://github.com/ruotianluo/DiscCaptioning}} \cite{disccap}, denoted as Att2in(D5) and FC(D5), 
where the retrieval reward weight is 5 (the CIDEr reward weight is 1), and likewise for D10;
(7) CVAE and GMMCVAE\footnote{\url{https://github.com/yiyang92/vae_captioning}} \cite{cvae}, (8) CGAN \cite{cgan}.

Models (1)-(7)  generate single caption for one image, and model (7) and (8) are able to generate diverse captions. \wqzn{The models are trained using Karpathy's training split}
\abcn{of MSCOCO}. 
We use each of the models to generate 10 captions for each image in the Karpathy's test split, which contains 5,000 images. The standard deviations of Gaussian noise \abcn{for GNC and DRV} are $\{1.0, 2.0, \cdots, 10.0\}$. 
For SS, we first generate a caption using beam search with beam-width 3, and then generate the other 9 captions by switching words with synonyms with probability $p\in\{0.1, 0.15, \cdots, 0.5\}$.  \abcn{Models and diversity generators are denoted as ``model-generator'', e.g., ``NIC-RS''.} 

The accuracy $acc$ of the generated captions $\calC$ is the average CIDEr: $\frac{1}{m}\sum_{i=1}^mCIDEr(c_i, \calC_{GT})$, where $c_i\in\calC$ and $\calC_{GT}$ is the set of human annotations. We also compute the {\em leave-one-out} accuracy of human annotations: $\frac{1}{N}\sum_{i=1}^NCIDEr(g_i, \calC_{GT\backslash i})$, where $g_i \in \calC_{GT}$ and $\calC_{GT\backslash i}$ is the set of human annotations without the i-th annotation. 
The diversity of $\calC$ is computed using the LSA-based method (denoted as LSA) and the kernel CIDEr method (denoted as Self-CIDEr), introduced in Sections \ref{LSA} and \ref{CIDEr}.
%
%
 Finally,  the accuracy and diversity tradeoff is summarized using the F-measure,  $F = \frac{(1+\beta^2)div\cdot acc}{\beta^2 div + acc}$.
 $\beta>1$ will weight accuracy  more, while $1>\beta\geq0$ will weight diversity more.

\section{Experiment Results}

We next present our experiment results evaluating methods based on both diversity and accuracy.

\subsection{Analysis of Caption Vocabulary}\label{GA} 
\begin{figure}[t]
\centering
\includegraphics[width=0.4\textwidth]{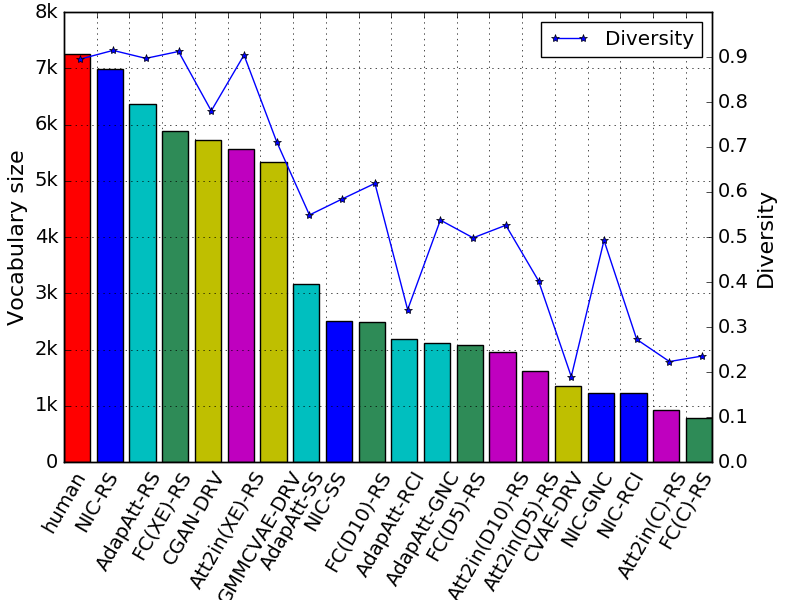}
\caption{The vocabulary sizes and diversity scores (Self-CIDEr)
of different caption models. The vocabulary of each trained model is collected from 50,000 captions (10 captions for each image), 
while the human annotations have 25,000 captions 
(5 captions for each image). For GNC, RCI, CVAE, GMMCVAE and CGAN, greedy search is used to generate captions. 
}\label{fig2}
\vspace{-1em}
\end{figure}

\begin{figure}[t]
\centering
\includegraphics[width=0.4\textwidth]{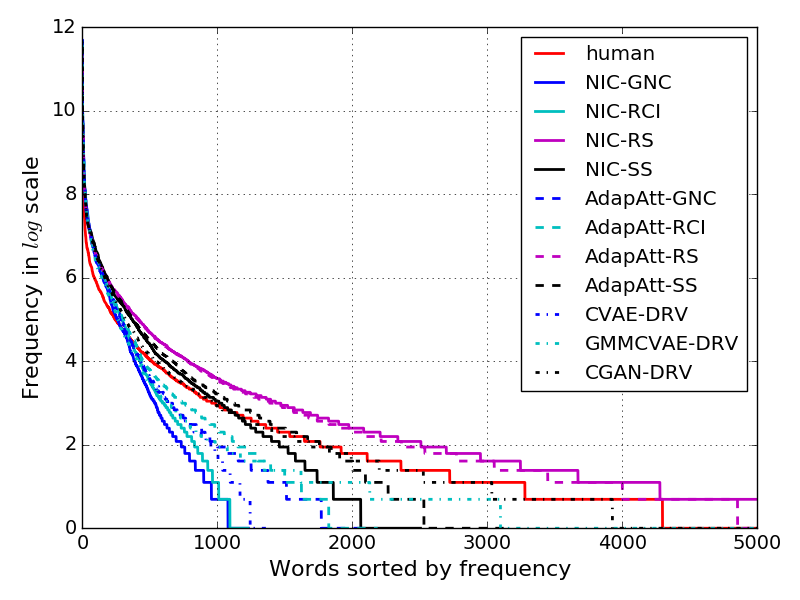}
\caption{Word frequency plots for the top 5,000 most-frequent words for each captioning model.
}\label{fig3}
\vspace{-1.5em}
\end{figure}

We first focus on the vocabulary of the generated captions from each model, including vocabulary size and word frequency. Generally, a large vocabulary size and long tail in the word frequency distribution indicates higher diversity.


Figure \ref{fig2} shows the vocabulary sizes of different models (here we only show the most representative models), as well as the models diversity score (Self-CIDEr).
Human annotations have the largest vocabulary, even though there are fewer human captions than model captions (25,000 for humans, and 50,000 for each model).
 For NIC and AdapAtt models, using RS results in larger vocabulary which also generates more diverse captions. 
Although AdapAtt is more advanced than NIC, the vocabulary size of AdapAtt-RS is smaller than that of NIC-RS.
\abcn{One possible reason is that models developed to obtain better accuracy metrics often learn to use more common words.}
Looking at reinforcement learning (e.g., Att2in(XE) vs. Att2in(C) vs. Att2in(D)), 
using CIDEr reward to fine-tune the model will drastically decrease the vocabulary size so as to improve the accuracy metric (CIDEr) 
\cite{scst}. 
Interestingly, using a retrieval reward gives a larger vocabulary size compared to using the CIDEr reward.
Improving retrieval reward encourages semantic similarity, while improving CIDEr reward encourages syntactic similarity, which leads to low diversity.
 Comparing the CGAN/CVAE methods, 
 CVAE has a smaller vocabulary compared to CGAN and GMMCVAE, 
which indicates that the latter could generate more diverse captions. Note that the vocabulary sizes only roughly reflects the diversity -- a small vocabulary could lead to diverse captions via using different combinations of words,
Hence, it is important to look at the pairwise similarity between captions.

Figure \ref{fig3} shows the frequency plots of each word used by the models. 
If a model employs diverse words, the plots in Figure \ref{fig3} should have a long tail. 
However, most of the models have learned to use around 2,000 common words.
In contrast, CGAN and GMMCVAE  encourage a longer-tail distribution, 
and in particular the word frequency plot of CGAN is similar to the human annotations. 
RS tends to give the most words, but also fails to generate fluent sentences.
Therefore, we suggest that both accuracy and diversity should be considered to evaluate a model. Interestingly, there is a very large gap between using cross-entropy and CIDEr rewards for reinforcement learning, which is bridged by the retrieval reward.
In Section \ref{RL}, we will show that 
 balancing cross-entropy, CIDEr, and retrieval rewards can also provide good results in terms of diversity and accuracy.

\subsection{Considering Diversity and Accuracy}\label{DA} 


\begin{figure*}[t]
\centering
\includegraphics[width=0.9\textwidth]{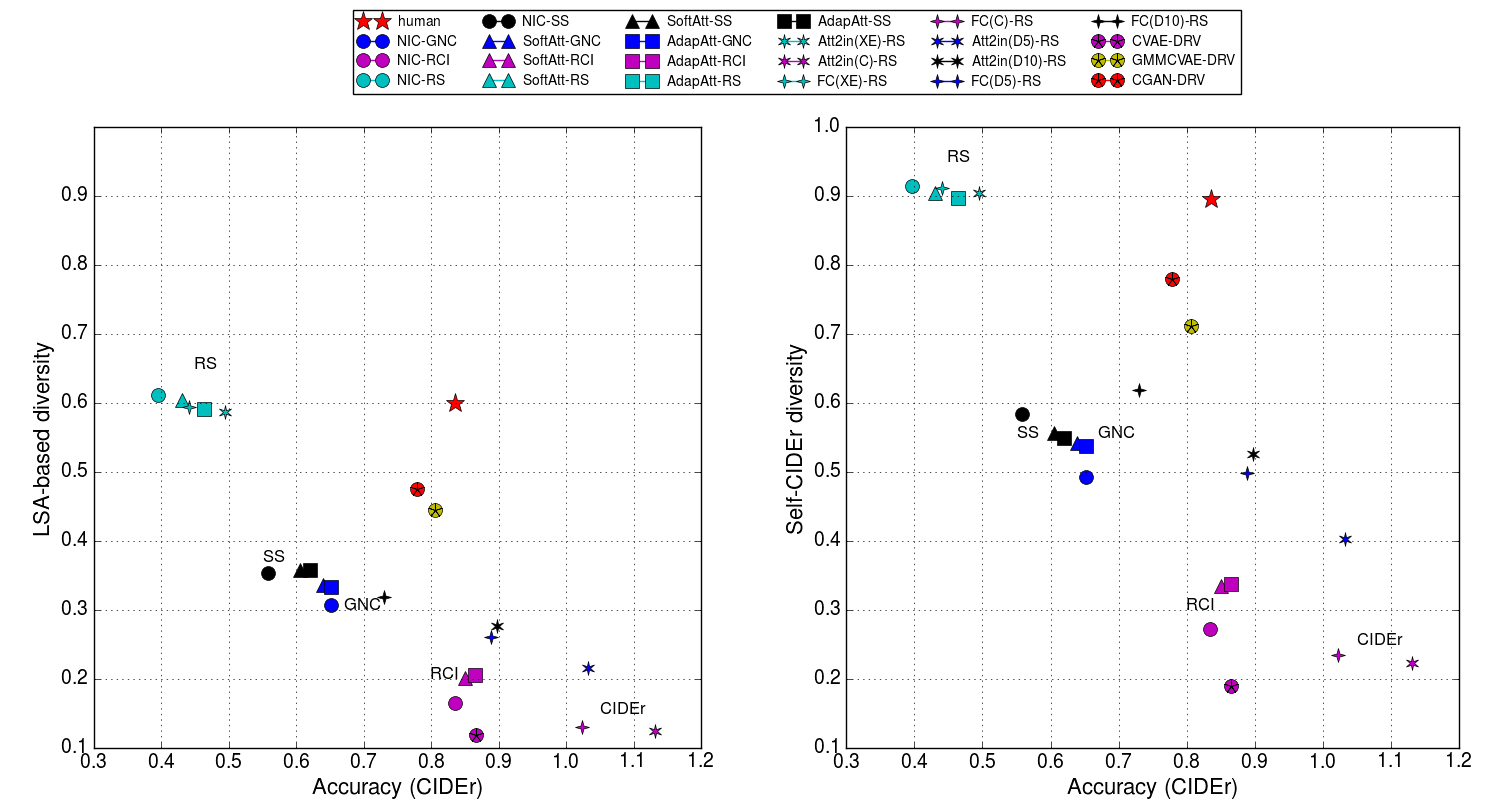}
\vspace{-1em}
\caption{The performance of different models considering accuracy and diversity. Left: using LSA-based diversity, which employs BoW features. Right: using CIDEr kernelized diversity (Self-CIDEr).
\abcn{The marker shape indicates the caption model, while the marker color indicates the diversity generator or training method.}
}\label{fig4}
\vspace{-1.5em}
\end{figure*}

\begin{figure}[t]
\centering
\includegraphics[width=0.45\textwidth]{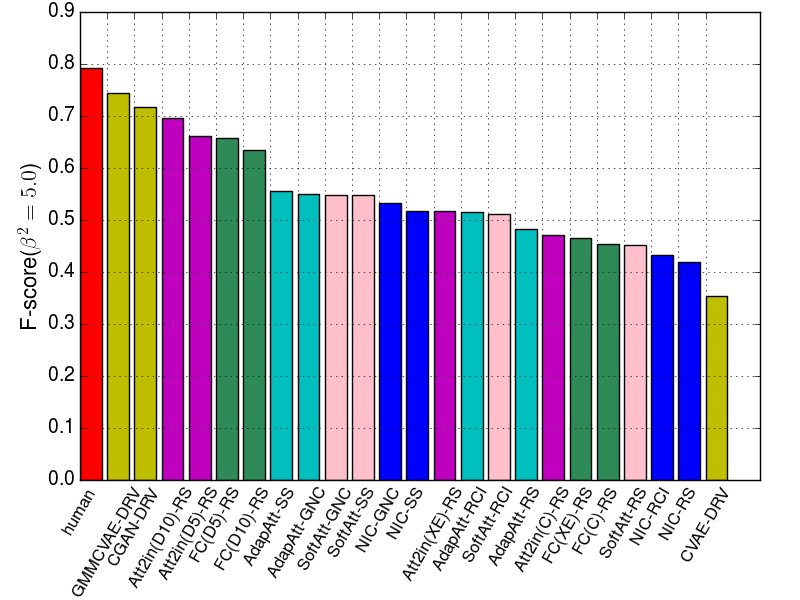}
\vspace{-1em}
\caption{The F-scores (using Self-CIDEr) of different models. 
}\label{fig5}
\vspace{-1.5em}
\end{figure}

\begin{figure}[t]
\centering
\includegraphics[width=0.45\textwidth]{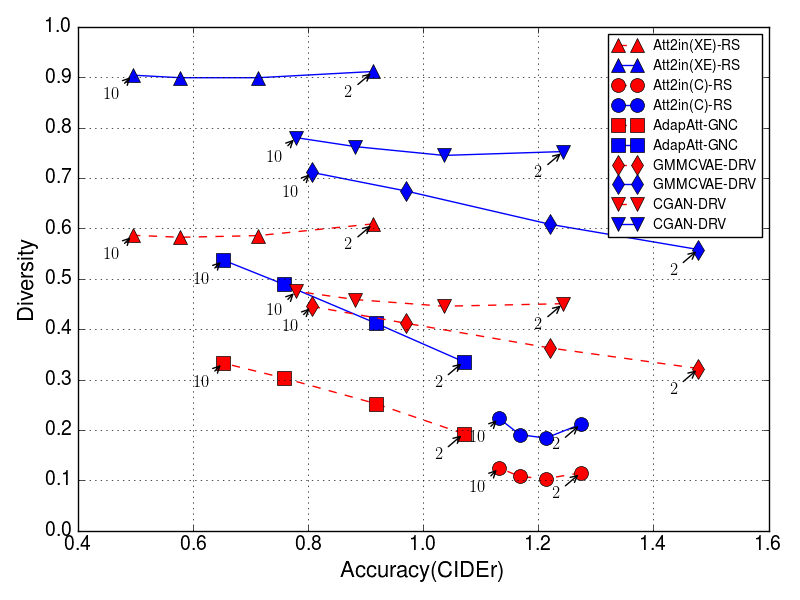}
\vspace{-1.5em}
\caption{The effects of different numbers of captions on diversity. We use $m\in\{2,5,8,10\}$. The dash lines represent LSA-based diversity and the solid lines represent Self-CIDEr diversity.}\label{fig6}
\vspace{-2em}
\end{figure}

\begin{figure*}[t!]
\centering
\begin{tabular}{@{}c@{}c@{}c@{}}
Self-CIDEr & LSA & mBLEU-mix\\
\includegraphics[width=0.33\textwidth]{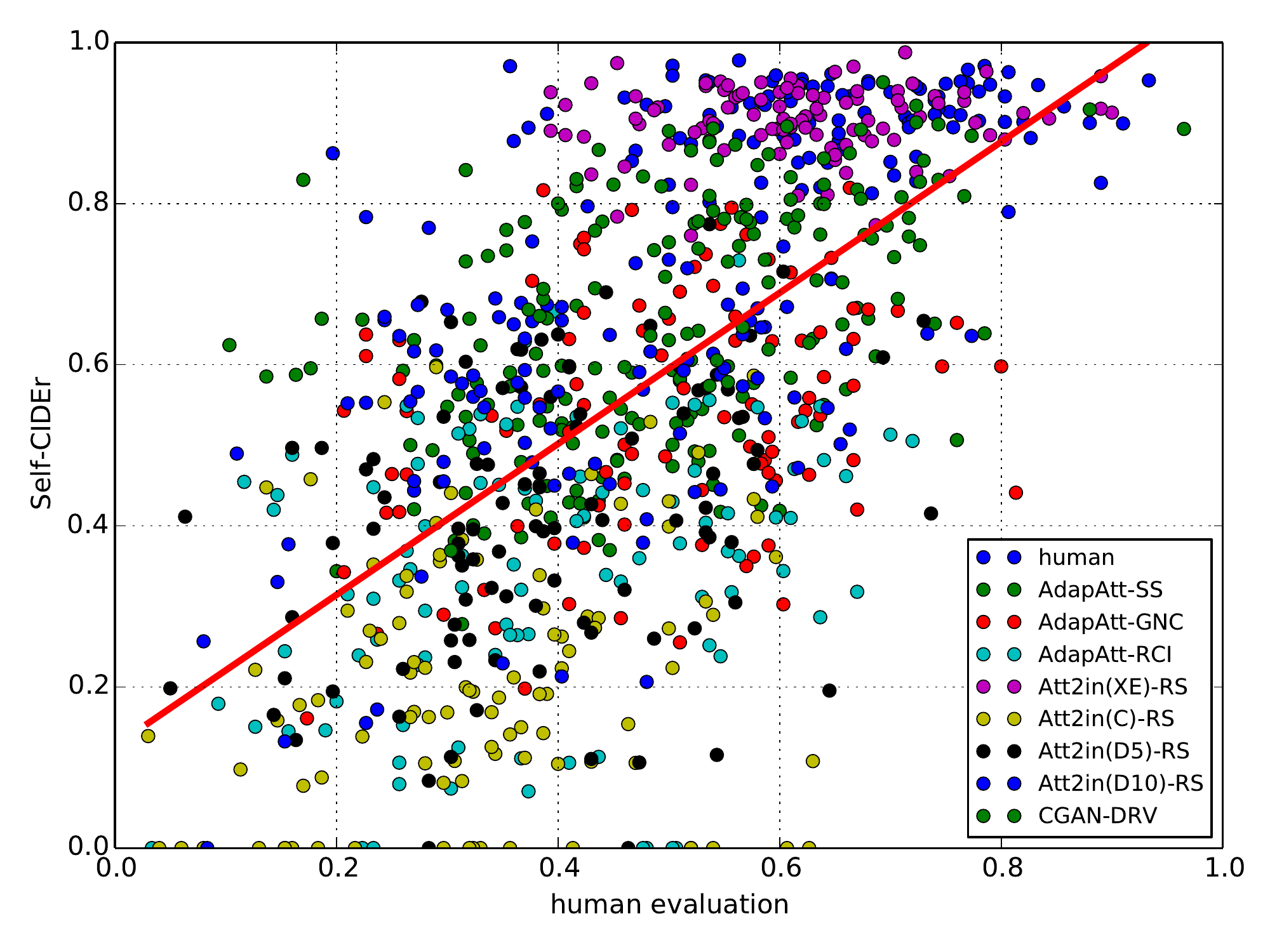}
&
\includegraphics[width=0.33\textwidth]{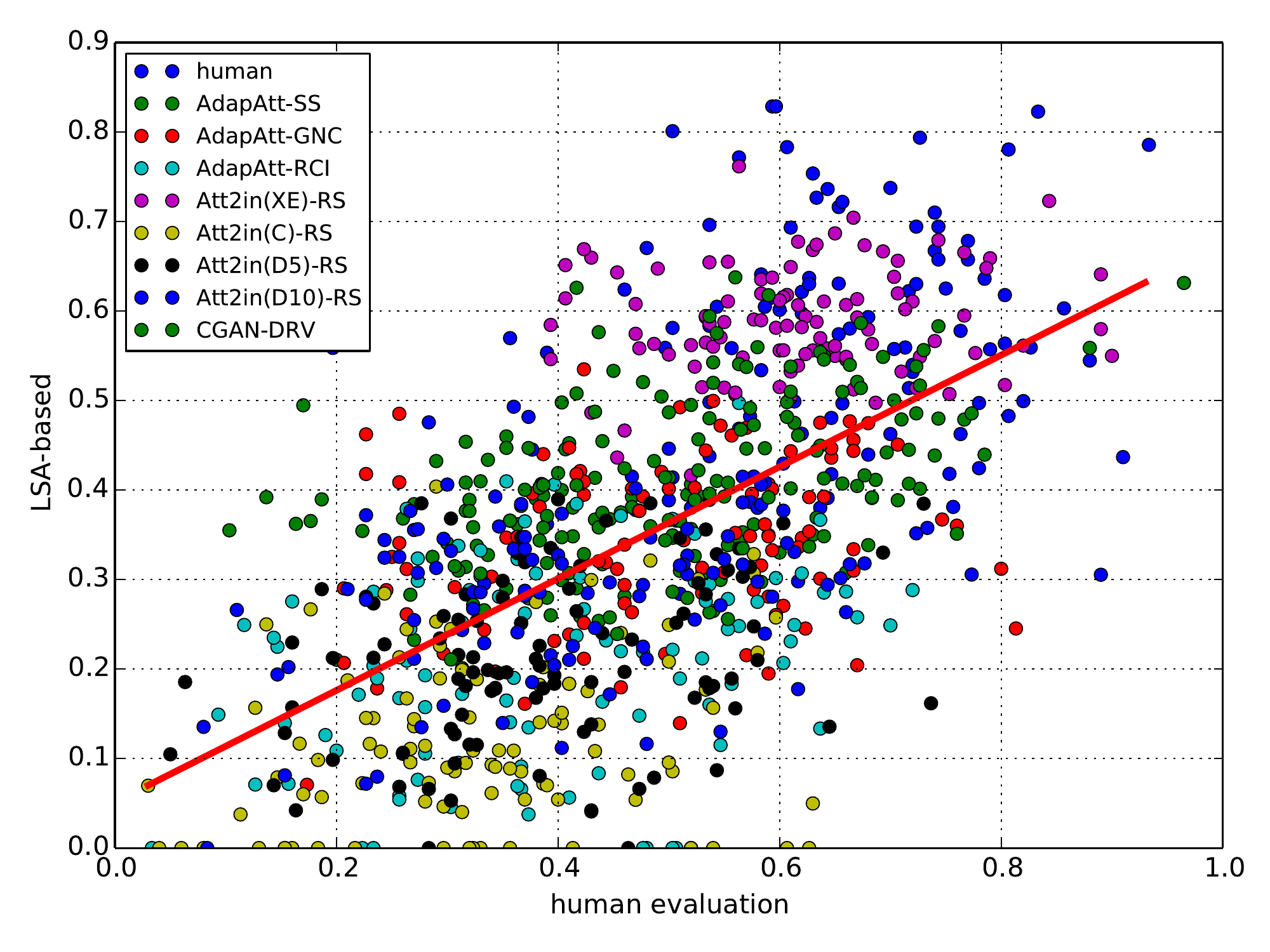}
&
\includegraphics[width=0.33\textwidth]{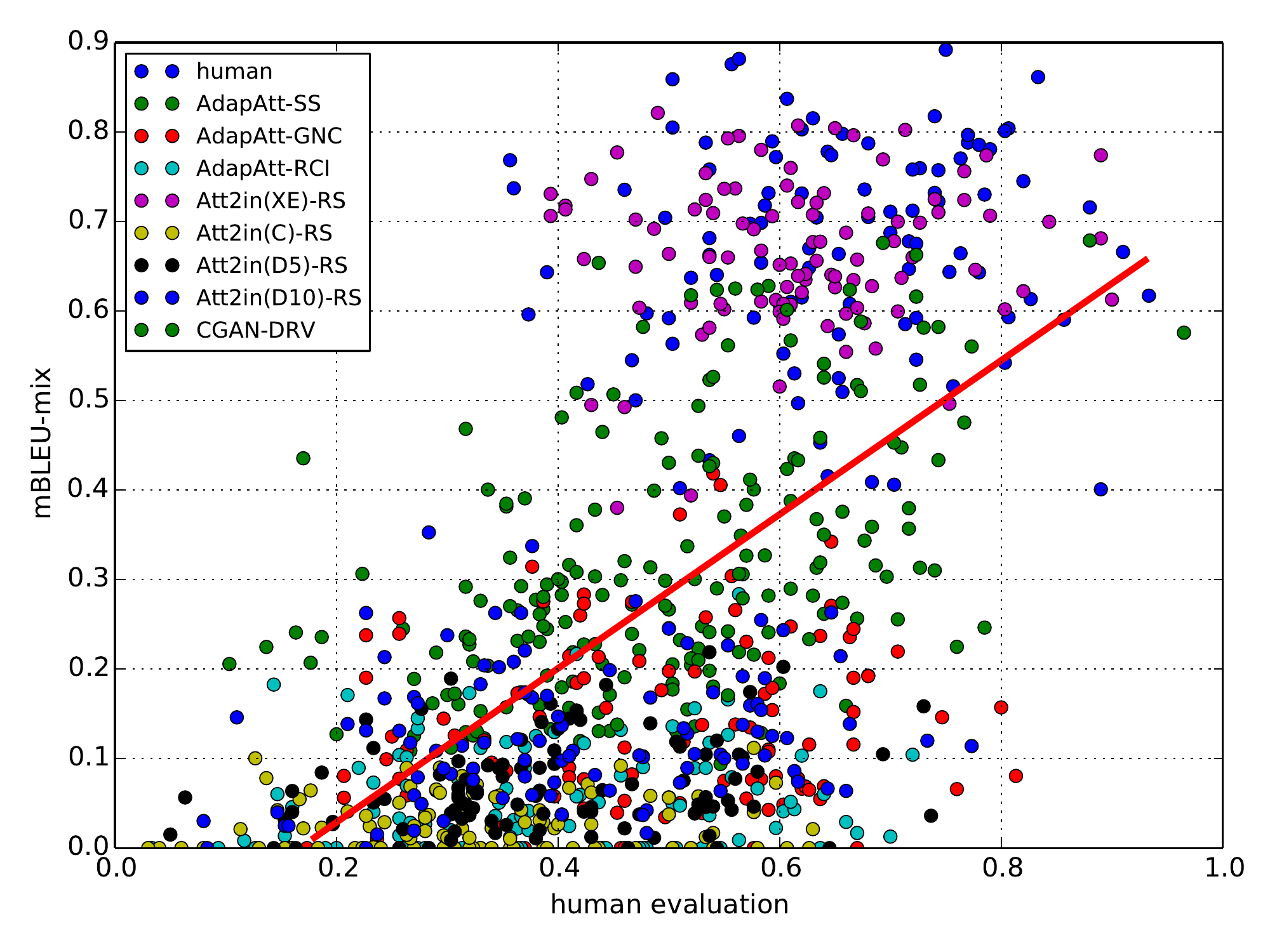}
\end{tabular}
\vspace{-1em}
\caption{The correlation plots between the diversity scores computed by different metrics and human evaluation. The red lines are the best fit lines to the data.
}\label{humaneval}
\vspace{-1em}
\end{figure*}

\begin{table}
\small
\centering
\begin{tabular}{@{}c|c@{\hspace{0.1cm}}c@{\hspace{0.1cm}}c@{}}
\hline
Corr Coef& Self-CIDEr & LSA & mBLEU-mix\\
\hline
overall Pearson $\rho$  & \bf 0.616 & 0.601 & 0.585\\
overall Spearman $\rho$& \bf 0.617 & 0.602 & 0.575  \\
\hline
avg. per image Spearman $\rho$& 0.674 &  \bf{0.678}  &  0.644 \\
\hline
\end{tabular}
\caption{Correlation  between computed diversity metric and human diversity judgement: (top) overall correlation; (bottom) correlation of per-image rankings of methods.
}
\label{corrs}
\vspace{-2.0em}
\end{table}

Here we re-evaluate the models accounting for both diversity and accuracy.
Figure \ref{fig4} shows the diversity-accuracy (DA) plots for LSA-based diversity and CIDEr kernelized diversity (Self-CIDEr). 
\abcn{The trends of LSA and Self-CIDEr are similar, although LSA yields overall lower values.}
Hence, we mainly discuss the results of Self-CIDEr.

 After considering both diversity and accuracy, we may need to rethink what should be considered a good model. 
We suggest that a good model should be close to human performance in the DA space. Looking at the performance of humans, the diversity is much higher than Att2in(C), which is considered a state-of-the-art captioning model. 
On the other hand, the diversity using randomly sampling (RS) 
are closer to human annotations.  However, the accuracy is poor, which indicates that the descriptions are not fluent or are off-topic. 
Therefore, a good model should well balance between diversity and accuracy. From this point of view, CGAN and GMMCVAE are among the best models, as they are closer to the human annotations in the DA space.
\abcnn{Example caption results and their diversity/accuracy metrics can be found in the supplemental.}

Most of the current state-of-the-art models are located in the bottom-right of the DA space, (high CIDEr score but poor diversity), as they aim to improve the accuracy.
For example, directly improving CIDEr reward via RL is a popular approach to obtain higher CIDEr scores \cite{scst,bmcr,disccap,disccap2}, but it encourages using common words and phrases (also see Figure \ref{fig2}), which lowers the diversity.
Using retrieval reward is able to improve diversity comparatively, e.g., Att2in(D5) vs Att2in(C),  because it encourages distinctive words and semantic similarity, and suppresses common syntaxes that do not benefit retrieval. The drawback of using retrieval model is that the fluency of the captions could be poor \cite{disccap}, and using a very large weight for the retrieval reward will cause the model to repeat the distinctive words. 
Finally, note that there is a large gap between using the cross-entropy loss and the CIDEr reward for training, e.g., Att2in(XE) and Att2in(C). \abcn{In the next subsection, we will consider building models to fill the performance gap by balancing between the losses.}

Comparing the diversity generators, SS and GNC are more promising for generating diverse captions. 
Captions generated using RCI  have higher accuracy, while those using RS  have higher diversity. Interestingly, in the top-left of the DA plot, using RS, a more advanced model can generate more accurate captions without reducing the diversity, 
This shows that an advanced model is able to learn a better $\hat{p}(c|I)$, which is more similar to the ground-truth distribution $p(c|I)$. \abcn{However, there is a long way to go to reach the accuracy of human annotations.}

\textbf{F-score comparison.}
Figure \ref{fig5} shows the F-scores that takes both diversity and accuracy into account. In this paper, we use $\beta^2=5.0$, which considers accuracy is more important than diversity. The reason for using a larger $\beta$ is that diverse captions \abcn{that do not describe the image well (low accuracy)} could be meaningless. 
The F-score of human annotations is the highest, and much higher than the models that generate only a single accurate caption for each image. 
CGAN and GMM-CVAE that are specifically designed to generate diverse captions also obtain high F-scores, which are closer to human performance, and this is consistent with Figure \ref{fig4}. For Att2in and FC, applying retrieval reward is a better way to achieve both diverse and accurate captions. Looking at the models using RS, more advanced models obtain higher F-score, which is also consistent with Figure \ref{fig4}. \wqzn{Note that the scales of diversity and accuracy scores are different.}

\textbf{The effects of number of captions.} 
We next consider how many captions should be generated to evaluate diversity.
Here, we use Att2in(XE)-RS, Att2in(C)-RS and AdapAtt-GNC, which are exemplar methods located in different areas 
in the DA plot (Fig.~\ref{fig4}), 
and GMMCVAE and CGAN, since they are the models with highest F-scores.
%
We first rank the captions for each image based on accuracy, then we select the top 2, 5, 8 and 10 captions to calculate the diversity scores (see Figure \ref{fig6}).

For Att2in(XE)-RS, which obtains very high diversity but low accuracy, the number of captions has small effect on the diversity. This is also seen in Att2in(C)-RS, which obtains low diversity but higher accuracy. 
The reason is that the captions generated by Att2in(XE)-RS are nearly always completely different with each other, 
\abcn{while captions generated by Att2in(C)-RS are almost always the same.} 
Therefore, increasing the number of generated captions does not affect the diversity for these models.
%
For models that well balance diversity and accuracy (e.g., CGAN  and GMMCVAE), more captions leads to higher diversity, and to some extent, diversity is linearly proportional to the number of captions. Therefore, we suggest that if a model is able to generate diverse captions, more captions should be generated to evaluate its diversity. Although the number of captions has an  effect on diversity scores, a better model generally obtains higher diversity.

\textbf{Correlation to human evaluation.} We conduct human evaluation on Amazon Machine Turk (AMT). We use 100 images, and for each image we show the worker 9 sets of captions, which are generated in different ways: human annotations and 8 models, AdapAtt-SS, AdapAtt-GNC, AdapAtt-RCI, Att2in(XE)-RS, Att2in(C)-RS, Att2in(D5)-RS, Att2in(D10)-RS and CGAN-DRV.
We require the workers to read all the sets of captions and then give scores (from 0 to 1) that reflects the diversity\footnote{In our instructions, diversity refers to different words, phrases, sentence structures, semantics or other factors that impact diversity.} of the set of captions. Each image is evaluated by 3 workers, and the diversity score for each image/model combination is the average score given by the 3 workers.

Fig.~\ref{humaneval} (left, center)
shows the correlation plots between our proposed metrics and human evaluation. The overall consistency between the proposed diversity metric and the human judgement is quantified using Pearson's (parametric) and Spearman's rank (non-parametric) correlation coefficients (see Table \ref{corrs} top). Since the human annotator evaluated the diversity scores for all methods on each image, they were implicitly ranking the diversity of the  methods. Hence, we also look at the consistency between the human rankings for an image and the rankings produced by the proposed metrics, as measured by the average per-image Spearman rank correlation (see Table \ref{corrs} bottom).
Both Self-CIDEr and LSA-based metrics are largely consistent with human evaluation of diveresity, with Self-CIDEr having higher overall correlation, while both have similar per-image ranking of methods.

We compare our metrics with $mBLEU_{mix}=1-\frac{1}{4}\sum_{n=1}^4 mBLEU_n$, which accounts for mBLEU-\{1,2,3,4\}, and we invert the score so that it is consistent with our diversity metrics (higher values indicate more diversity).
The correlation plot between mBLEU-mix and human judgement is shown in Fig.~\ref{humaneval} (right). mBLEU-mix has lower correlation coefficient with human judgement, compared to LSA and Self-CIDEr (see Table \ref{corrs}). Similar results are obtained when looking at the mBLEU-n scores.
Self-CIDEr has better overall correlation with human judgement, while the two methods are comparable in terms of per-image consistency of method ranking.

In addition the correlation plot shows the mBLEU scale is not uniformly varying, with more points falling at the lower and higher ends of the scale and less points in the middle. 
In contrast, LSA and Self-CIDEr have more uniform scales.
Note that another advantage of our proposed metrics is that they can be used to analyze the latent semantics and visualize the caption's diversity (see our supplementary material), which mBLEU cannot do.

\begin{figure*}[t]
\centering
\includegraphics[width=0.49\textwidth]{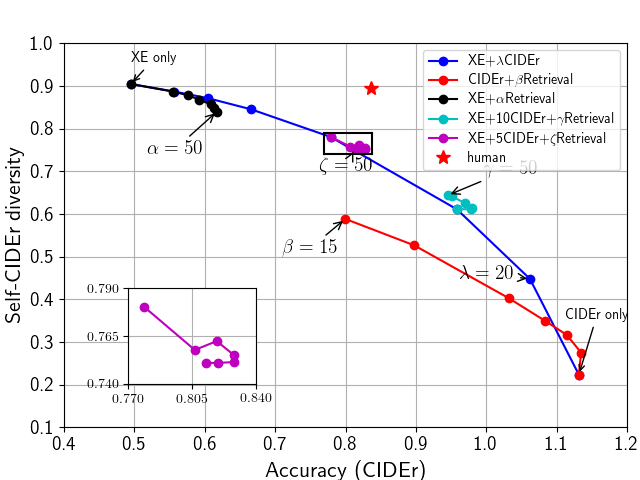}
\includegraphics[width=0.49\textwidth]{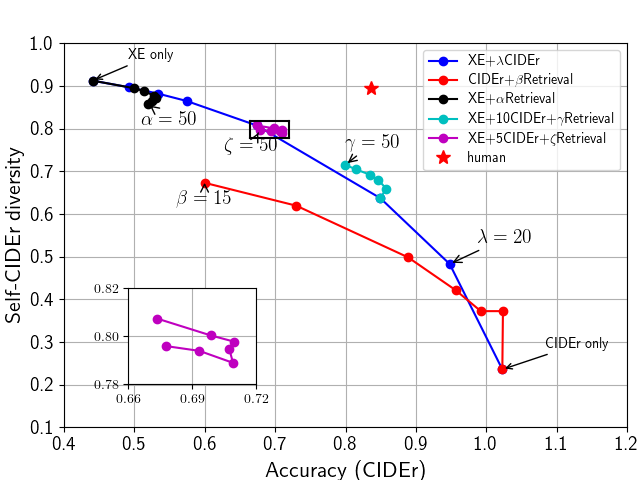}
\vspace{-0.5em}
\caption{The diversity and accuracy performance of Att2in (left) and FC (right) with different loss functions. XE, CIDEr, and Retrieval denote the cross-entropy loss, CIDEr reward \cite{scst} and retrieval reward \cite{disccap}.  The weights are $\lambda\in\{0,1,2,3,5,10,20\}$, $\beta\in\{0, 1, 2, 3, 5, 10, 15\}$, $\alpha\in\{0, 5, 10, 20, 30, 40, 50\}$, $\gamma\in\{0, 10, 20, 30, 40, 50\}$ and $\zeta\in\{0, 5, 10, 20, 30, 40, 50\}$.
\abcnn{The inset plot in the bottom-left is a zoom-in of rectangle in the main plot.}
}\label{fig7}
\vspace{-1em}
\end{figure*}

\begin{figure*}[t]
\centering
\footnotesize
\begin{tabular}{@{}c@{}c@{}c@{}c@{}c@{}}
XE-only & $\lambda=5$ & $\lambda=10$ &  $\lambda=20$ & CIDEr-only 
\\
\includegraphics[width=0.19\textwidth]{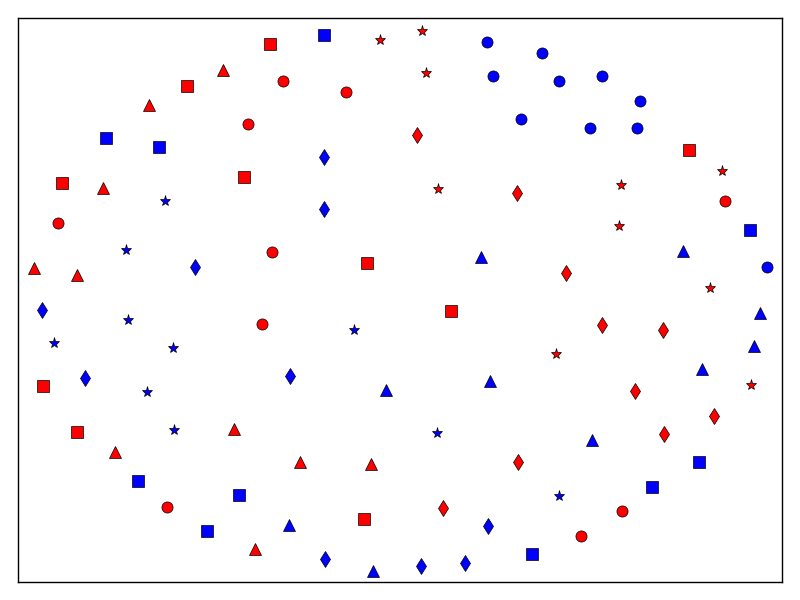}
&
\includegraphics[width=0.19\textwidth]{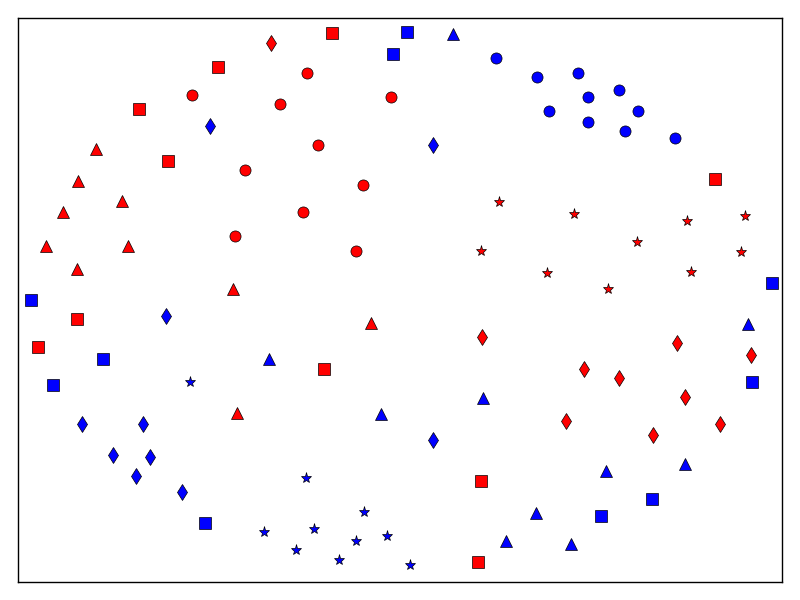}
&
\includegraphics[width=0.19\textwidth]{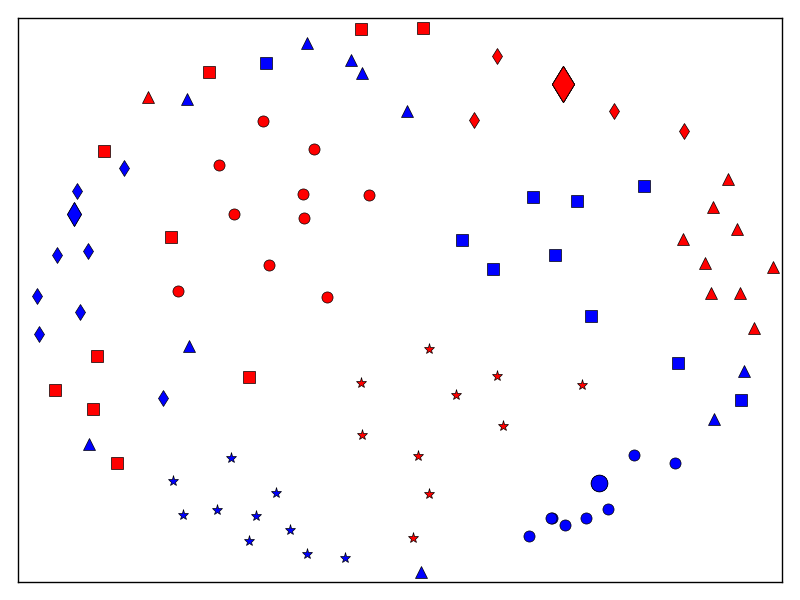}
&
\includegraphics[width=0.19\textwidth]{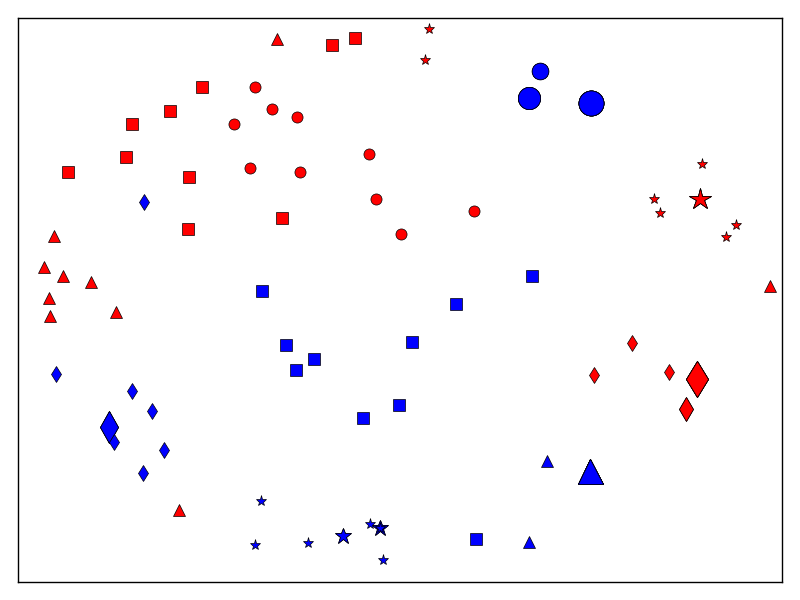}
&
\includegraphics[width=0.19\textwidth]{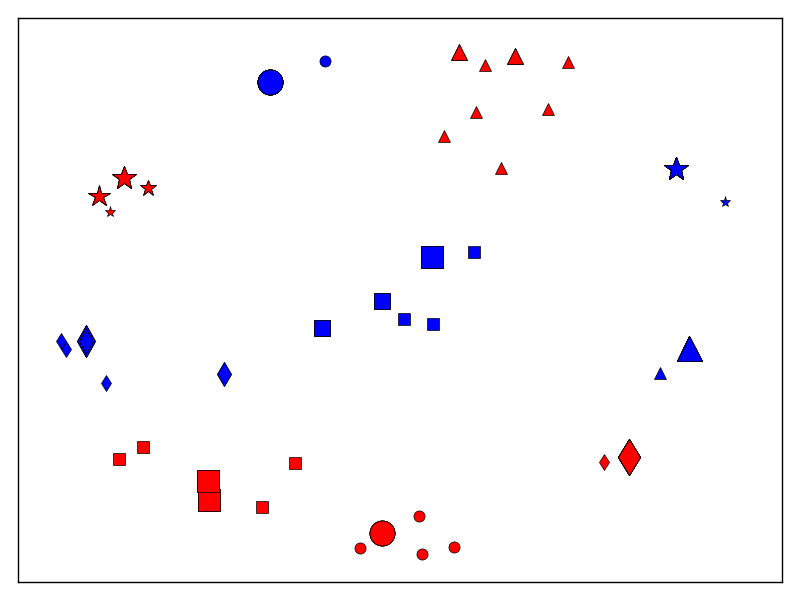}
\end{tabular}
\vspace{-1em}
\caption{MDS visualization of the similarity between captions of 10 images. 10 captions are randomly sampled from Att2in for each image.
\abcn{Markers and colors indicate different images.}
Larger
markers indicates multiple captions located at the same position.}\label{fig8}
\vspace{-1.5em}
\end{figure*}

\subsection{Re-thinking Reinforcement Learning}\label{RL} 

\CUT{
\begin{figure*}[t]
\centering
\includegraphics[width=0.45\textwidth]{figures/f1.png}
\includegraphics[width=0.45\textwidth]{figures/f5.png}
\caption{F-scores of Att2in and FC models trained with different loss functions. Left: F1-scores, right: F5-scores.}\label{fig9}
\end{figure*}
}

In this subsection we further investigate RL for image captioning,
\abcn{and how to bridge the gap between the performance using cross-entropy and using CIDEr reward.
In particular, we train Att2in and FC using different loss functions that combine the cross-entropy, CIDEr reward, and retrieval reward, with varying weights.
}

The results are shown in Figure \ref{fig7}.
Balancing the XE loss and CIDEr reward is the most effective way to bridge the gap.
Using larger weight $\lambda$ results in 
higher accuracy but lower diversity. Based on our experiments, using $\lambda=5$
well balances diversity and accuracy, resulting in performance that is closer to the human annotations, \abcnn{and similar to CGAN and GMMCVAE}.
%
Using XE loss only, 
the learned distribution $\hat{p}(c|I)$ has a large variance, which could be very flat and smooth, and thus incorrect words appear during sampling. In contrast, using CIDEr reward can suppress the probability of the words that cannot benefit CIDEr score, and encourage the words that improve CIDEr.  \abcn{Hence, combining the two losses  suppresses the poor words and promotes good words (CIDEr), while also preventing the distribution from concentrating to a single point (XE).}
\abcn{Figure \ref{fig8} visualizes the similarity between captions using multi-dimensional scaling (MDS) \cite{MDS}, for different values of $\lambda$.  As $\lambda$ increases, some captions are repeated, and points are merged in the MDS visualization.}


Finally, using the retrieval reward in the combined loss function
also slightly improves the diversity and accuracy, and generally results in a local move in the DA plot.
However, a very large $\gamma$ or $\zeta$ could result in a repetition problem, \ie, a model will repeat the distinctive words, since distinctive words are more crucial for the retrieval reward. \wqznn{We show more examples of generated captions in our supplementary material.}
\vspace{-0.5em}
\section{Conclusion}
\vspace{-0.5em}
\abcn{In this paper, we have developed a new metric for evaluating the diversity of a caption set generated for an image.  Our diversity measure is based on computing singular values (or eigenvalues) of the kernel matrix composed of CIDEr values between all pairs of captions, which is interpretable as performing LSA on the weighted n-gram feature representation to extract the topic-structure of the captions.}
%
Using our diversity metric and CIDEr to re-evaluate recent captioning models, we found that: 1) models that have optimized accuracy tend to have very low diversity, and there is a large gap between model and human performances;
2) balancing the XE loss and other reward functions when using RL is a promising way to generate diverse and accurate captions, which can achieve performance that is on par with generative models (CGAN and GMMCVAE).


\appendix \section{Supplemental}
\subsection{Comparison with mBLEU}

\begin{figure*}[t]
\centering
\includegraphics[width=0.9\textwidth]{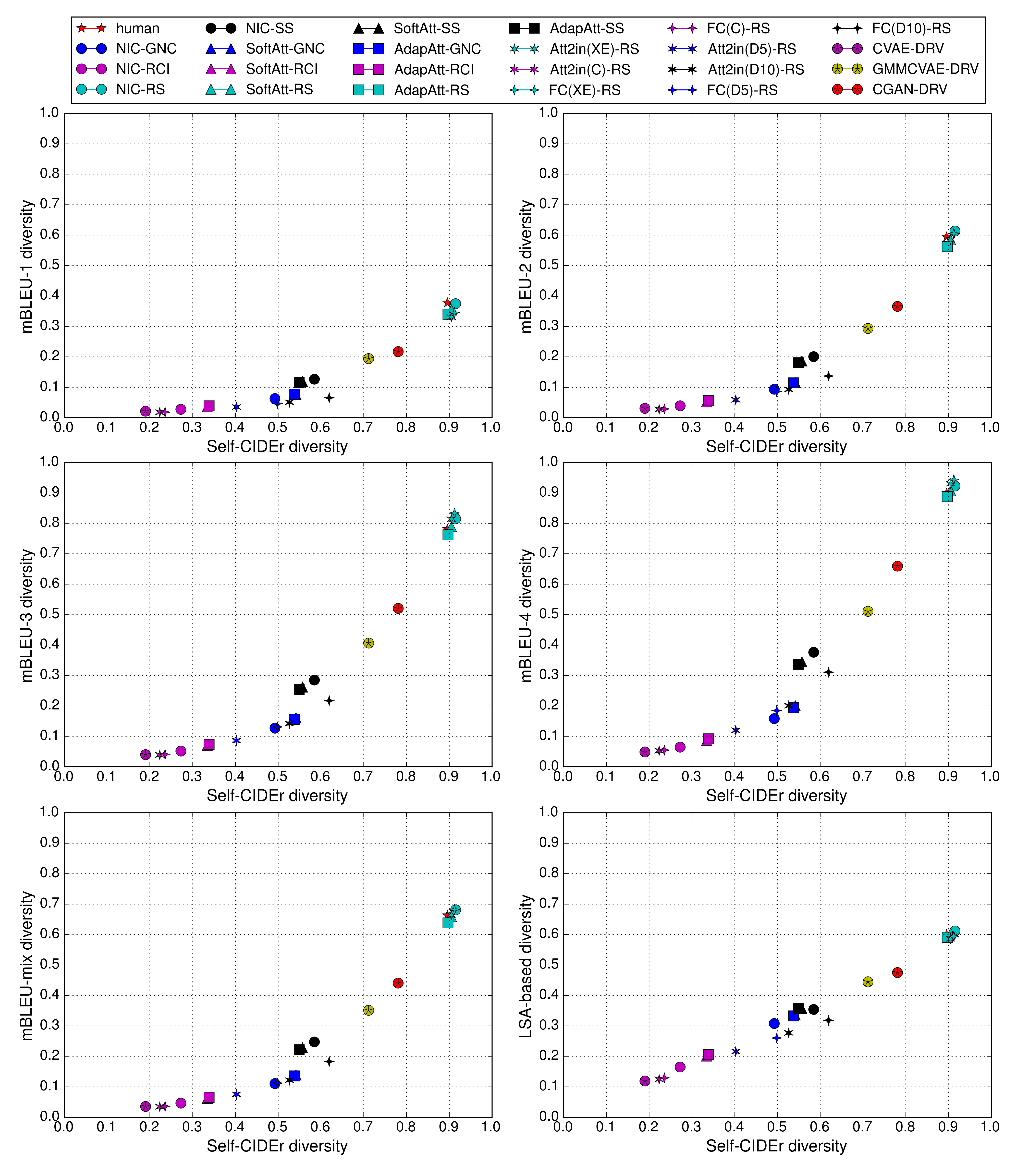}
\caption{The correlation between our Self-CIDEr diversity and mBLEU scores. We use 5 mBLEU scores---uni-, bi-, tri-, quad- and mixed BLEU scores. To some extent, our Self-CIDEr diversity score is consistent with mBLEU diversity scores, and it shows an exponential correlation between mBLEU diversity scores and Self-CIDEr diversity score. The reason is that in Self-CIDEr we use $-\log_m(r)$ as the final diversity score. The most different is that our Self-CIDEr assigns a higher diversity score to FC(D10)-RS, by contrast the mBLEU metric assigns a higher score to NIC-SS (see the ranking of different models in figure \ref{supp-fig1-rank}). Recall that SS approach applies synonyms to replace the words in a captions, which just changes the words but could not change the semantics, and Self-CIDEr diversity metric that is derived from latent semantic analysis (LSA) pays much attention to \textit{semantic diversity}, hence, using synonyms could result in low Self-CIDEr diversity. Moreover, mBLEU-1,2,3,4 and mix could assign differnt rankings to the same model, \eg, FC(D5)-RS is ranked below NIC-GNC using mBLEU-1,2, whereas mBLEU-4 assigns a higer score to FC(D5)-RS than NIC-GNC, by contrast, both of them obtain similar diversity score using mBLEU-3,mix and Self-CIDEr. The correction between LSA-based and Self-CIDEr is roughly linear.}\label{supp-fig1}
\end{figure*}

\begin{figure*}[t]
\centering
\includegraphics[width=\textwidth]{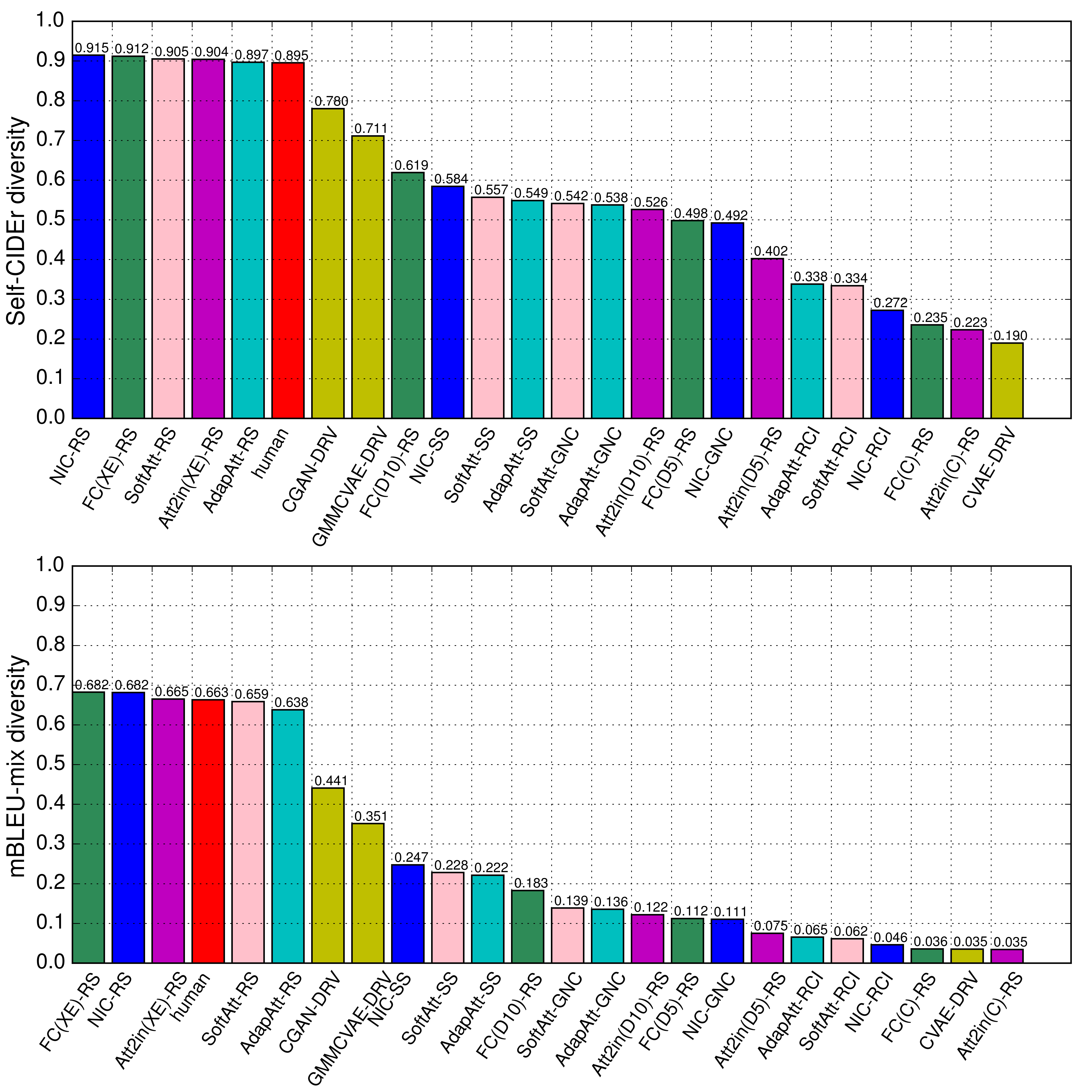}
\caption{Model ranking by Self-CIDEr diversity and the mixed mBLEU diversity. Generally, mBLEU-mix diversity scores are lower than Self-CIDEr diversity scores, although both of them account for uni-, bi-, tri- and quad-grams. In most cases, the two diversity metrics provide consistent ranking, except for the rankings of human and FC(D10)-RS. Human is ranked below SoftAtt-RS and AdapAtt-RS using Self-CIDEr , while mBLEU-mix ranks human above them. Another difference is that FC(D10)-RS is treated as a more diverse model than the SS models using Self-CIDEr diversity metric, in contrast, using mBLEU-mix, FC(D10)-RS obtains a lower diversity score.}\label{supp-fig1-rank}
\end{figure*}

\begin{figure*}[t]
\centering
\includegraphics[width=0.9\textwidth]{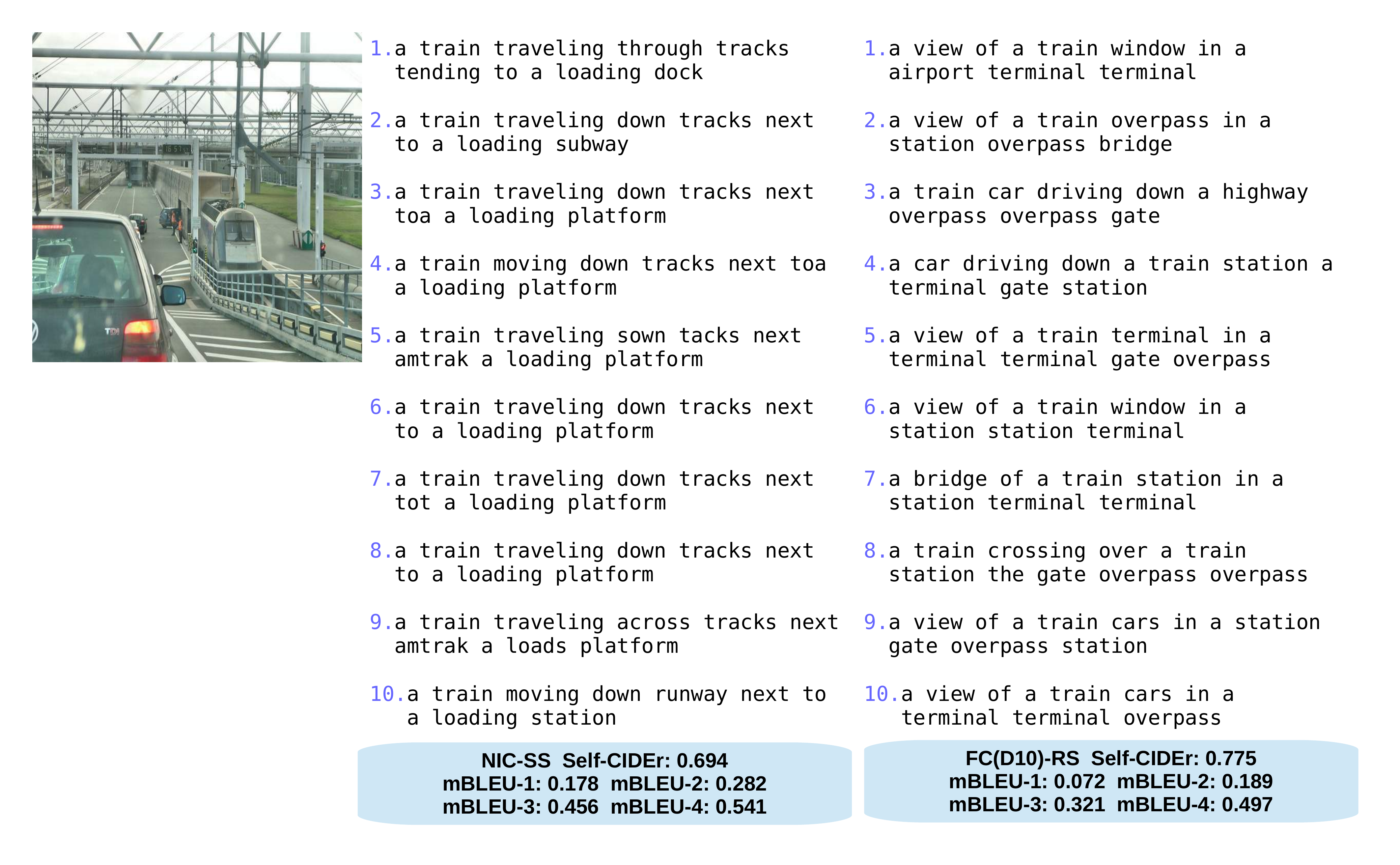} \\
\vspace{-1.5em}
\includegraphics[width=0.9\textwidth]{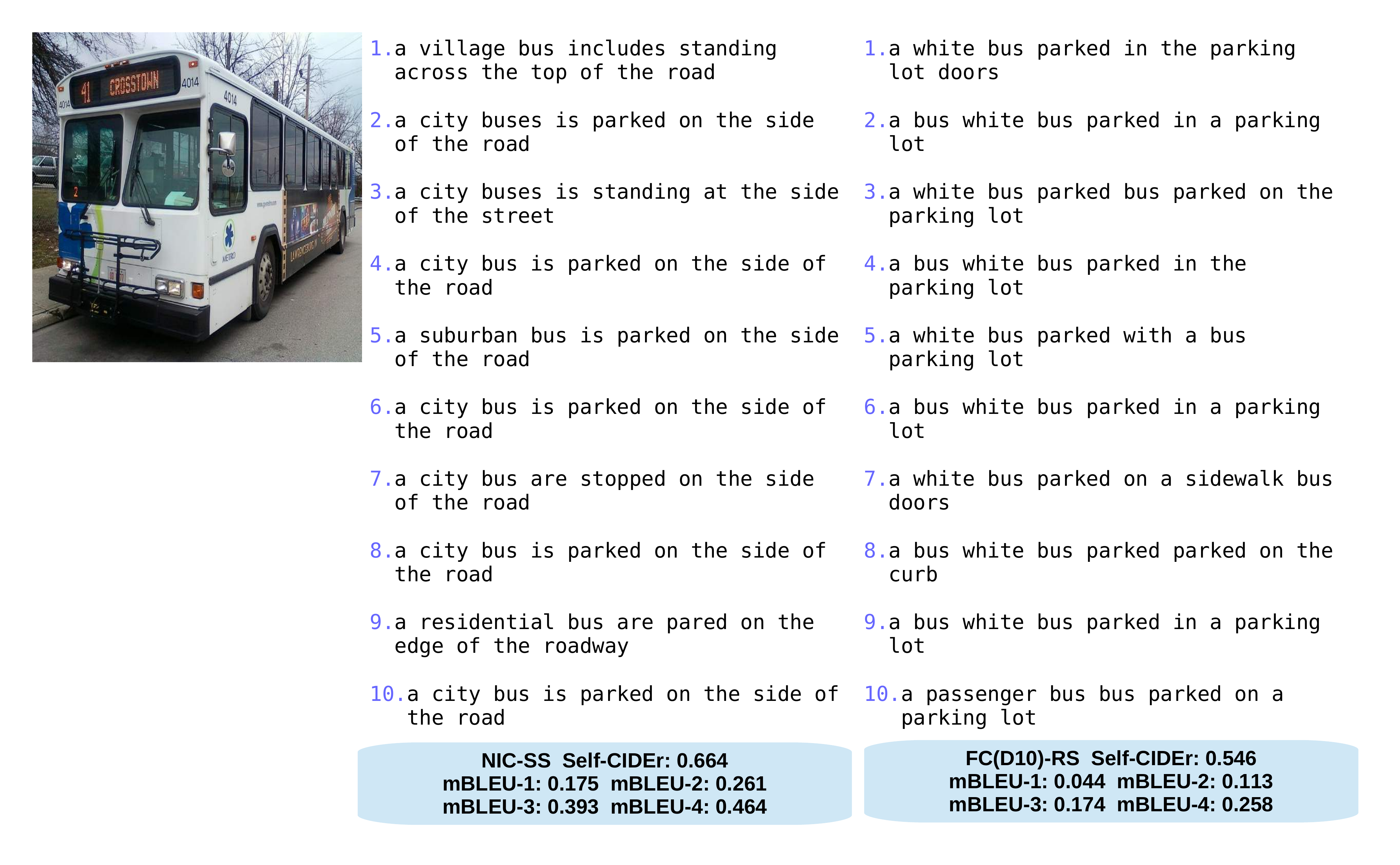}
\vspace{-1.5em}
\caption{Generated captions of NIC-SS and FC(D10)-RS models and the corresponding diversity scores and a high score indicates diverse captions. For the first image (top), Self-CIDEr and mBLEU metrics provide different rankings of NIC-SS and FC(D10)-RS. NIC-SS is ranked below FC(D10)-RS based on Self-CIDEr, while the mBLEU diversity scores of NIC-SS are higher. Looking at the captions, NIC-SS just switches ``traveling'' to ``moving'', ``down'' to ``through'' or ``across'' and ``platform'' to ``station'', while FC(D10)-RS describes differnt concepts, such as ``train'', ``car(s)'', ``airport terminal'', ``stations'', ``overpass'' and ``bridge'', therefore, FC(D10)-RS obtains higher Self-CIDEr diversity score but lower mBLEU diversity scores. In terms of the second image (bottom), Self-CIDEr and mBLEU divsity metrics provide consitent ranking, because both NIC-SS and FC(D10)-RS describes the same thing---\textit{bus parked on}. Comparing the two images, the first one contains more concepts than the second one, and using SS just changes the words but does not change the semantics, in contrast, FC(D10)-RS introduces different concepts to different captions, which could result in differnt semantics, however, FC(D10)-RS could generate influent sentences.}\label{supp-fig2}
\end{figure*}

\begin{figure*}[t]
\centering
\includegraphics[width=0.95\textwidth]{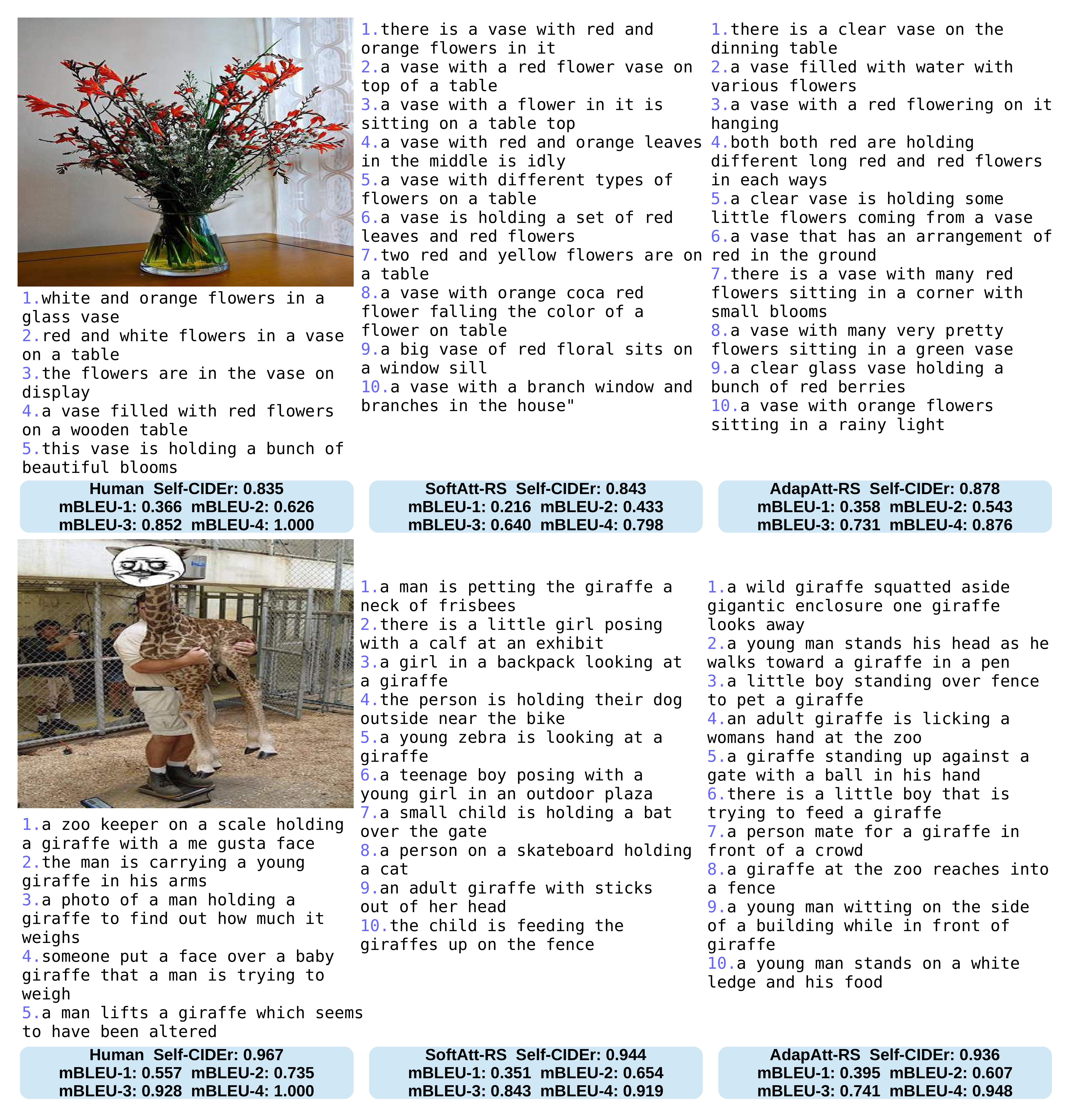}
\caption{Captions generated by human, SoftAtt-RS and AdaptAtt-RS and the corresponding diversity scores and a high score indicates diverse captions. For the first image (top), human annotations obtain the lowest Self-CIDEr diversity score but highest mBLEU diversity scores. Looking at the captions, although human annotations use different words and phrases, they describe the same concept---\textit{vase with flowers on a table}, whereas SoftAtt-RS describes not only \textit{vase}, \textit{flowers} and \textit{table}, but also \textit{leaves}, \textit{window} and \textit{house}, and AdapAtt-RS uses ``clear'', ``water'', ``rainy light'' and ``green'' to describe the image, which could be plausible descriptions. For the second image, both Self-CIDEr and mBLEU diversity metrics provide high scores and all captions almost use different words. Note that we use $-\log$ function in Self-CIDEr diversity metric, which could be flat and less sensitive if the captions are relatively diverse (see figure \ref{supp-fig1}).}\label{supp-fig2-2}
\end{figure*}

Given a set of captions $\calC=\{c_1, c_2, \cdots, c_m\}$, mBLEU \cite{cgan1} is computed as follows:
\begin{equation}\label{supp-eq1}
mBLEU_n = \frac{1}{m}\sum_{i=1}^m BLEU_n(c_i, \calC\backslash i),
\end{equation}
where $n$ represents using n-gram, $BLEU_n()$ represents the BLEU function and $\calC\backslash i$ denotes the set of captions without the i-th caption. A higher mBLEU score indicates lower diversity, here we use $1-mBLEU_n$ to measure the diversity of $\calC$, thus, a higher score indicates higher diversity. We also consider the mixed mBLEU score, which is the weighted sum of $mBLEU_n$, \ie, $\sum_{n=1}^4\omega_n mBLEU_n$, and in this paper we set $\omega_n=\frac{1}{4}$.

Figure \ref{supp-fig1} shows the correlation between our Self-CIDEr diversity metric and the mBLEU diversity metric. Figure \ref{supp-fig1-rank} shows the rankings of different models based on Self-CIDEr diversity and mBLEU-mix diversity.

Figure \ref{supp-fig2} shows the captions generated by NIC-SS and FC(D10)-RS and the corresponding diversity scores, and figure \ref{supp-fig2-2} shows the sets of captions generated by human, SoftAtt-RS and AdapAtt-RS and the corresponding diversity scores. 

Another advantage of using LSA-based and Self-CIDEr diversity metrics is that we can project the captions into a latent semantic space vis decomposition, thus we are able to visualize each set of captions in the space composed of 2 latent semantics (see section \ref{supp-vis}).

\subsection{Visualizing Captions via Decomposition} \label{supp-vis}

\begin{figure*}[t]
\centering
\includegraphics[width=\textwidth]{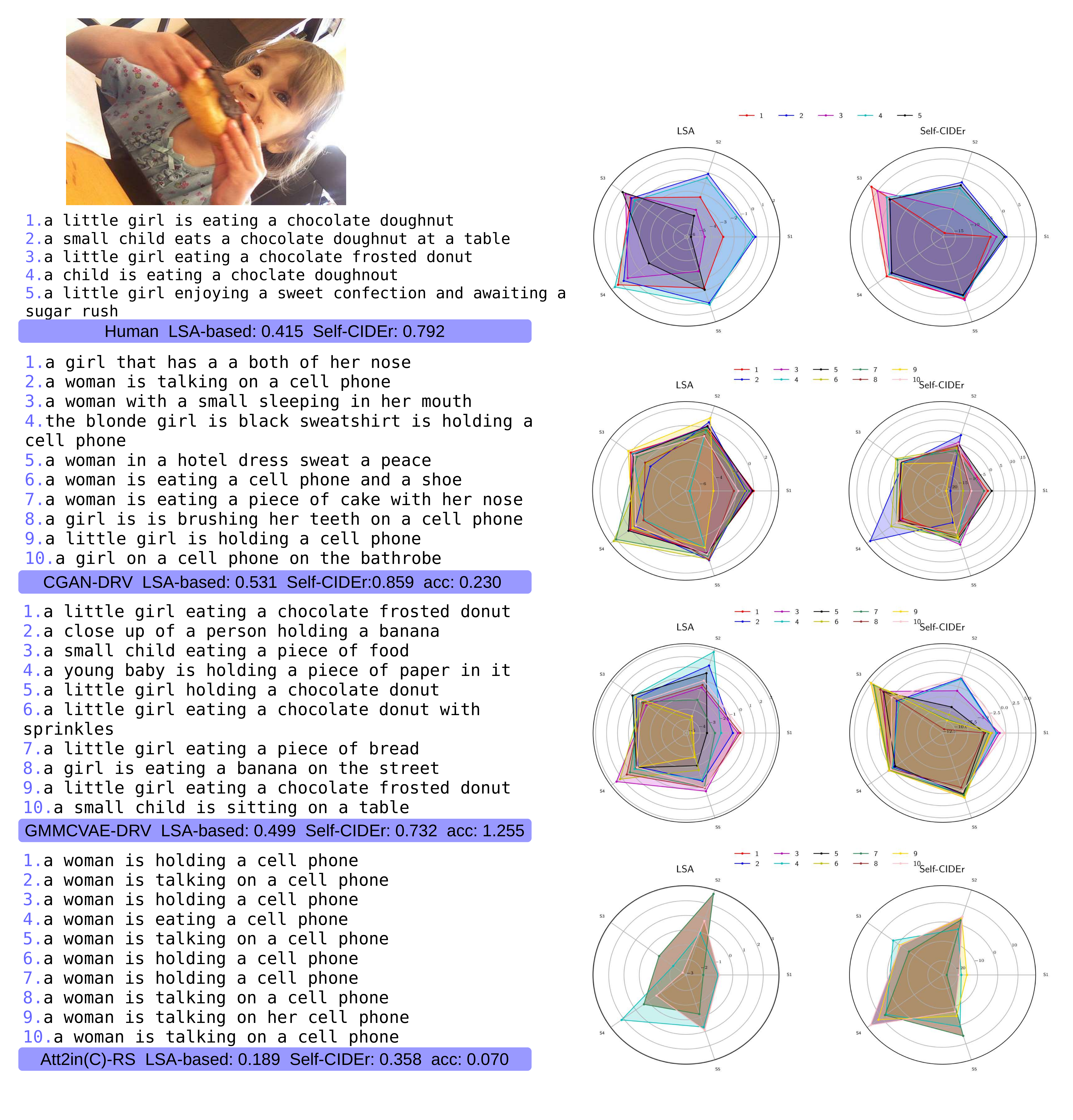}
\vspace{-2em}
\caption{Visualizing captions in the latent semantic space. Human annotations focus on ``little girl'', ``eating'' and ``chocolate donut'', and looking at the radar chart of LSA, it roughly contains 2 semantics---S3 and S4. Moreover, captions 3 and 5 talk more about S3, and caption 5 does not talk about S4, if we compare captions 3 and 5, it is easy to find that both of them use ``a little girl'', but caption 3 also uses ``eating'' and ``doughnout'', therefore, S4 could denote ``eating'' somthing. Captions 6, 7 of CGAN, captions 3, 6 of GMMCVAE and caption 4 that use ``eating'' have a larger value of S4. Similarly, in the radar chart of Self-CIDEr, S3 could represent ``eating'' somthing and S4 could denote ``talking'', thus the captions contain ``eating'' have relative large values of S3 and the captions that use ``talking'' could have larger values of S4.}\label{supp-fig3}
\end{figure*}

\begin{figure*}[t]
\centering
\includegraphics[width=\textwidth]{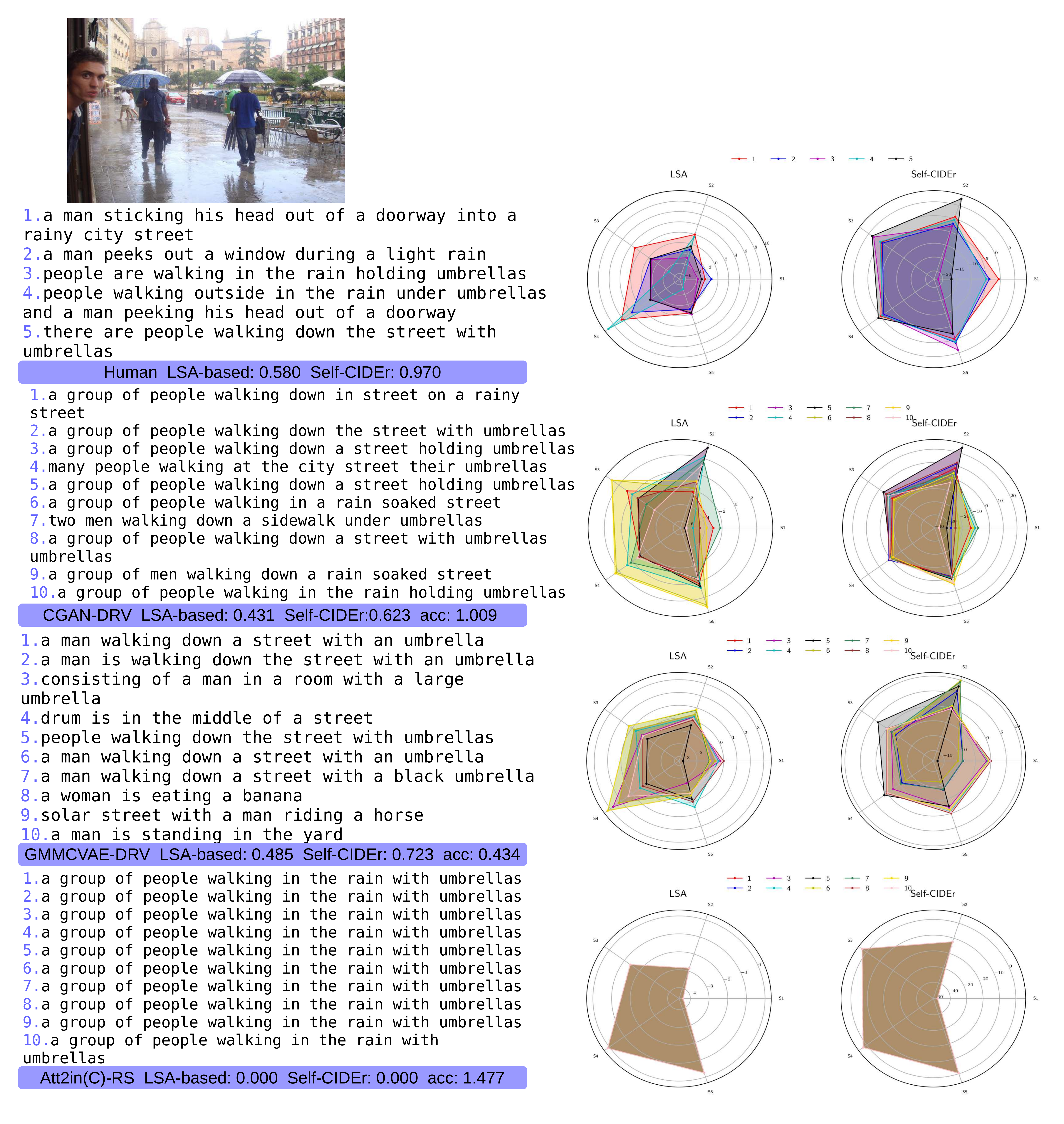}
\caption{Visualizing captions in the latent semantic space.}\label{supp-fig4}
\end{figure*}

\begin{figure*}[t]
\centering
\includegraphics[width=\textwidth]{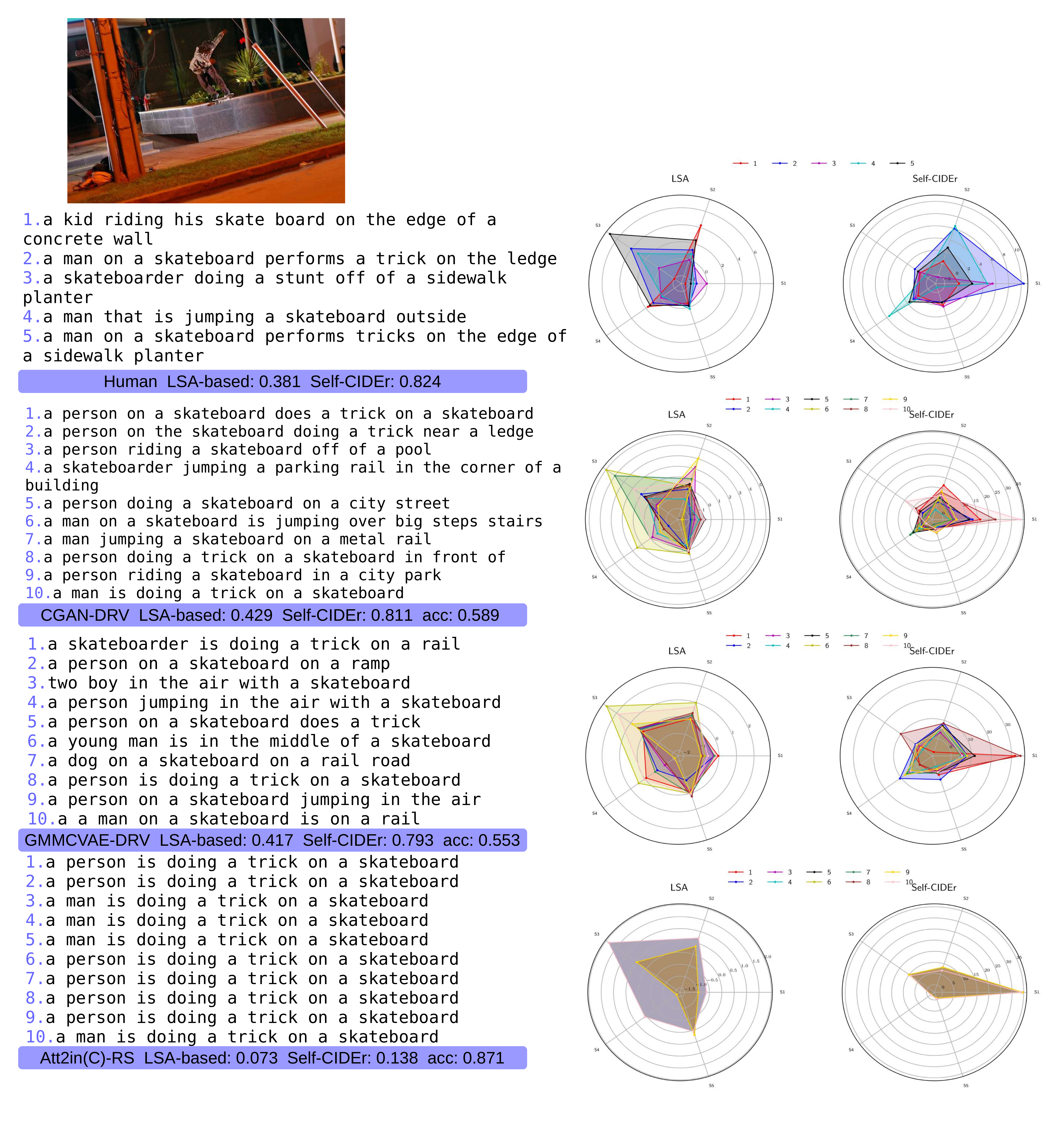}
\caption{Visualizing captions in the latent semantic space.}\label{supp-fig5}
\end{figure*}


In this section we visualize the captions using LSA and our proposed kernelized method (Self-CIDEr) to project captions into the semantic space, thus we can see what the captions are talking about. Given a set of captions $\calC$, we first construct a dictionary $\calD$ and then each caption is represented by bag-of-word features, and $\calC$ is represented by the ``word-caption'' matrix $\mathbf{M}$ (we have described the details in our paper). To better visualize the captions, we use stop words in LSA\footnote{\url{https://www.nltk.org/book/ch02.html}}.

Recall that $\mathbf{M}=\mathbf{U}\mathbf{S}\mathbf{V}^T$ using SVD and we select the 5 largest singular values and their corresponding row vectors $[\mathbf{v}_1; \mathbf{v}_2; \cdots; \mathbf{v}_5]$ in $\mathbf{V}^T$, where $\mathbf{v}_i\in \mathbb{R}^m$, $m$ denotes the number of captions. In LSA, $\mathbf{v}_i$ reflects the relationship between captions and the i-th latent semantic. Hence, the j-th caption $c_j$ can be represented by a 5-D vector $[v_1^j, v_2^j, \cdots, v_5^j]$, where $v_i^j$ is the j-th elements of $\mathbf{v}_i$.  Similarly, for the kernelized method, we can decompose the kernel matrix $\mathbf{K}$, thus $\mathbf{K}=\mathbf{V}\mathbf{\Lambda}\mathbf{V}^T$ and the 5-D vector $[v_1^j, v_2^j, \cdots, v_5^j]$ can also represent a caption in the semantic space. In this paper we use radar charts to illustrate the correlations between captions and latent semantics. A positive value could represent that a caption contains this semantic, a negative or null value could indicate that a caption does not contain this semantic and it describes another thing. Figure \ref{fig3} to \ref{fig5} show the captions generated by different models and each caption in the 5-D latent semantic space. Note that the semantics in LSA and Self-CIDEr could be different. 

\subsection{Captions Generated by RL-based Methods}

\begin{figure*}[t]
\centering
\includegraphics[width=\textwidth]{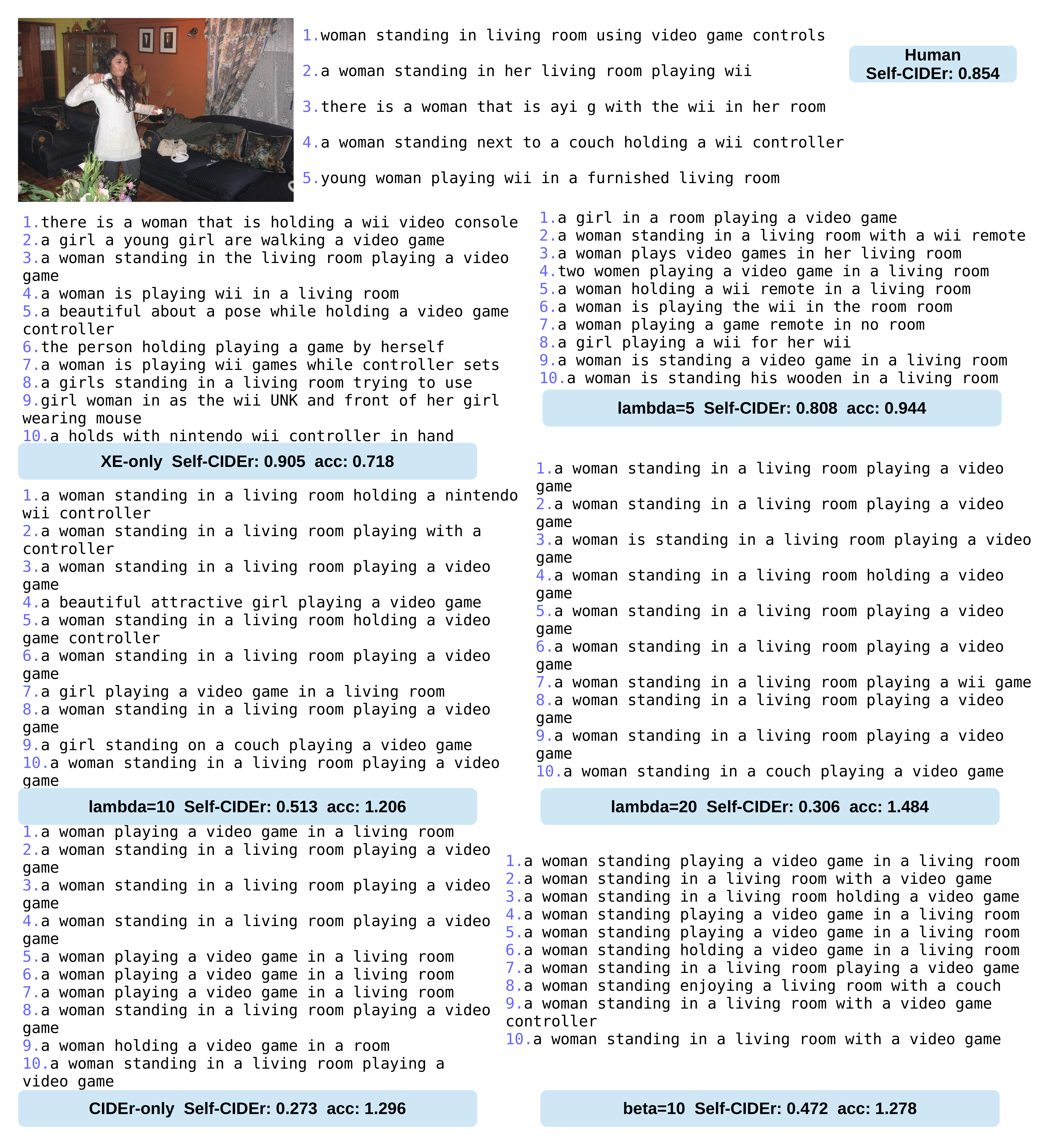}
\caption{Captions generated by Att2in model traind with different conbinations of loss functions. \textbf{lambda} denotes the weight of CIDEr reward and \textbf{beta} denotes the weight of retrieval reward (see section 5.3 in our paper). Obviously, using large \textbf{lambda} is able to increase the accuracy but reduce diversity.}\label{supp-fig8}
\end{figure*}

\begin{figure*}[t]
\centering
\includegraphics[width=\textwidth]{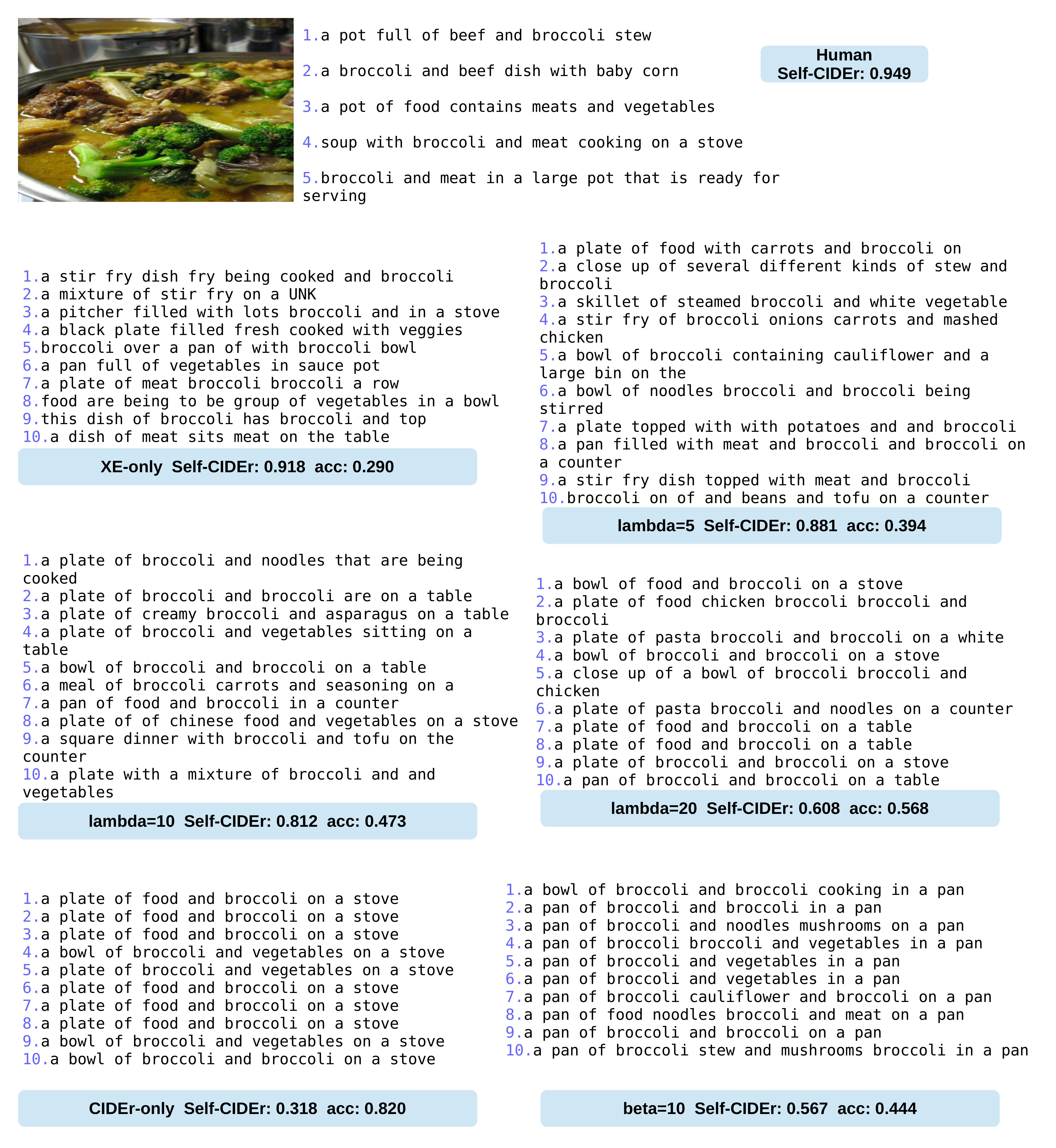}
\caption{Captions generated by Att2in model traind with different conbinations of loss functions. \textbf{lambda} denotes the weight of CIDEr reward and \textbf{beta} denotes the weight of retrieval reward (see section 5.3 in our paper).}\label{supp-fig9}
\end{figure*}

We show some generated captions of RL-based methods with different combinations of loss functions (see figure \ref{supp-fig8} and \ref{supp-fig9}).

{\small
\bibliographystyle{ieee_fullname}
\bibliography{myref}

\begin{thebibliography}{10}\itemsep=-1pt

\bibitem{spice}
Peter Anderson, Basura Fernando, Mark Johnson, and Stephen Gould.
\newblock Spice: Semantic propositional image caption evaluation.
\newblock In {\em ECCV}, 2016.

\bibitem{convimagecap}
Jyoti Aneja, Aditya Deshpande, and Alexander Schwing.
\newblock Convolutional image captioning.
\newblock In {\em CVPR}, 2018.

\bibitem{MDS}
Andreas Buja, Deborah~F Swayne, Michael~L Littman, Nathaniel Dean, Heike
  Hofmann, and Lisha Chen.
\newblock Data visualization with multidimensional scaling.
\newblock {\em Journal of Computational and Graphical Statistics},
  17(2):444--472, 2008.

\bibitem{learn2eval}
Yin Cui, Guandao Yang, Andreas Veit, Xun Huang, and Serge Belongie.
\newblock Learning to evaluate image captioning.
\newblock In {\em CVPR}, 2018.

\bibitem{cgan}
Bo Dai, Sanja Fidler, Raquel Urtasun, and Dahua Lin.
\newblock Towards diverse and natural image descriptions via a conditional gan.
\newblock In {\em ICCV}, 2017.

\bibitem{lsa}
S. Deerwester, S. Dumais, G. Furnas, T. Landauer, and R. Harshman.
\newblock Indexing by latent semantic analysis.
\newblock {\em Journal of the American Society for Information Science},
  41:391--407, 1990.

\bibitem{M}
M. Denkowski and A. Lavie.
\newblock Meteor universal: Language specific translation evaluation for any
  target language.
\newblock In {\em EACL Workshop on Statistical Machine Translation}, 2014.

\bibitem{visconcept}
Hao Fang, Saurabh Gupta, Forrest Iandola, Rupesh~K. Srivastava, Li Deng, Piotr
  Dollar, Jianfeng Gao, Xiaodong He, Margaret Mitchell, John~C. Platt, C.
  Lawrence~Zitnick, and Geoffrey Zweig.
\newblock From captions to visual concepts and back.
\newblock In {\em CVPR}, 2015.

\bibitem{templatemodel}
Ali Farhadi, Mohsen Hejrati, Mohammad~Amin Sadeghi, Peter Young, Cyrus
  Rashtchian, Julia Hockenmaier, and David Forsyth.
\newblock Speaking the same language: Matching machine to human captions by
  adversarial training.
\newblock In {\em ECCV}, 2010.

\bibitem{scn}
Zhe Gan, Chuang Gan, Xiaodong He, Yunchen Pu, Kenneth Tran, Jianfeng Gao,
  Lawrence Carin, and Li Deng.
\newblock Semantic compositional networks for visual captioning.
\newblock In {\em CVPR}, 2017.

\bibitem{scenegraph}
Justin Johnson, Ranjay Krishna, Michael Stark, Li-Jia Li, David~Ayman Shamma,
  Michael Bernstein, and Li Fei-Fei.
\newblock Image retrieval using scene graphs.
\newblock In {\em CVPR}, 2015.

\bibitem{evalWMD}
Mert Kilickaya, Aykut Erdem, Nazli Ikizler-Cinbis, and Erkut Erdem.
\newblock Re-evaluating automatic metrics for image captioning.
\newblock In {\em EACL}, 2017.

\bibitem{babytalk}
G. Kulkarni, V. Premraj, V. Ordonez, S. Dhar, S. Li, Y. Choi, A.~C. Berg, and
  T.~L. Berg.
\newblock Babytalk: Understanding and generating simple image descriptions.
\newblock {\em IEEE Transactions on Pattern Analysis and Machine Intelligence},
  35(12):2891--2903, 2013.

\bibitem{WMD}
Matt Kusner, Yu Sun, Nicholas Kolkin, and Kilian Weinberger.
\newblock From word embeddings to document distances.
\newblock In {\em ICML}, 2015.

\bibitem{ngrammodel}
S. Li, G. Kulkarni, T.~L. Berg, A.~C. Berg, and Y. Choi.
\newblock Composing simple image descriptions using web-scale n-grams.
\newblock In {\em CoNLL}, 2011.

\bibitem{R}
C.-Y. Lin.
\newblock Rouge: A package for automatic evaluation of summaries.
\newblock In {\em ACL Workshop}, 2004.

\bibitem{bmcr}
Siqi Liu, Zhenhai Zhu, Ning Ye, Sergio Guadarrama, and Kevin Murphy.
\newblock Improved image captioning via policy gradient optimization of spider.
\newblock In {\em ICCV}, 2017.

\bibitem{disccap2}
Xihui Liu, Hongsheng Li, Jing Shao, Dapeng Chen, and Xiaogang Wang.
\newblock Show, tell and discriminate: Image captioning by self-retrieval with
  partially labeled data.
\newblock In {\em ECCV}, 2018.

\bibitem{when2look}
Jiasen Lu, Caiming Xiong, Devi Parikh, and Richard Socher.
\newblock Knowing when to look: Adaptive attention via a visual sentinel for
  image captioning.
\newblock In {\em CVPR}, 2017.

\bibitem{disccap}
Ruotian Luo, Brian Price, Scott Cohen, and Gregory Shakhnarovich.
\newblock Discriminability objective for training descriptive captions.
\newblock In {\em CVPR}, 2018.

\bibitem{word2vec}
Tomas Mikolov, Ilya Sutskever, Kai Chen, Greg~S Corrado, and Jeff Dean.
\newblock Distributed representations of words and phrases and their
  compositionality.
\newblock In {\em NIPS}, 2013.

\bibitem{bleu}
K. Papineni, S. Roukos, T. Ward, and W.-J. Zhu.
\newblock Bleu: a method for automatic evaluation of machine translation.
\newblock In {\em ACL}, 2002.

\bibitem{scst}
Steven~J Rennie, Etienne Marcheret, Youssef Mroueh, Jerret Ross, and Vaibhava
  Goel.
\newblock Self-critical sequence training for image captioning.
\newblock In {\em CVPR}, 2017.

\bibitem{text2scenegraph}
Sebastian Schuster, Ranjay Krishna, Angel Chang, Li Fei-Fei, and Christopher~D.
  Manning.
\newblock Generating semantically precise scene graphs from textual
  descriptions for improved image retrieval.
\newblock In {\em EMNLP-Vision and Language Workshop}, 2015.

\bibitem{kernelbook}
John Shawe-Taylor and Nello Cristianini.
\newblock {\em Kernel Methods for Pattern Analysis}.
\newblock Cambridge University Press, 2004.

\bibitem{cgan1}
Rakshith Shetty, Marcus Rohrbach, and Lisa~Anne Hendricks.
\newblock Speaking the same language: Matching machine to human captions by
  adversarial training.
\newblock In {\em ICCV}, 2017.

\bibitem{vggnet}
Karen Simonyan and Andrew Zisserman.
\newblock Very deep convolutional networks for large-scale image recognition.
\newblock {\em arXiv preprint arXiv:1409.1556}, 2014.

\bibitem{C}
R. Vedantam, C.~Lawrence Zitnick, and D. Parikh.
\newblock Cider: Consensus-based image description evaluation.
\newblock In {\em CVPR}, 2015.

\bibitem{NIC}
Oriol Vinyals, Alexander Toshev, Samy Bengio, and Dumitru Erhan.
\newblock Show and tell: A neural image caption generator.
\newblock In {\em CVPR}, 2015.

\bibitem{cvae}
Liwei Wang, Alexander Schwing, and Svetlana Lazebnik.
\newblock Diverse and accurate image description using a variational
  auto-encoder with an additive gaussian encoding space.
\newblock In {\em NIPS}, 2017.

\bibitem{cnnpluscnn}
Qingzhong Wang and Antoni~B. Chan.
\newblock Cnn+cnn: Convolutional decoders for image captioning.
\newblock {\em arXiv preprint arXiv:1805.09019}, 2018.

\bibitem{gha}
Qingzhong Wang and Antoni~B. Chan.
\newblock Gated hierarchical attention for image captioning.
\newblock In {\em ACCV}, 2018.

\bibitem{attributes}
Qi Wu, Chunhua Shen, Lingqiao Liu, Anthony Dick, and Anton van~den Hengel.
\newblock What value do explicit high level concepts have in vision to language
  problems?
\newblock In {\em CVPR}, 2016.

\bibitem{spatt}
Kelvin Xu, Jimmy Ba, Ryan Kiros, Kyunghyun Cho, Aaron Courville, Ruslan
  Salakhudinov, Rich Zemel, and Yoshua Bengio.
\newblock Show, attend and tell: Neural image caption generation with visual
  attention.
\newblock In {\em ICML}, 2015.

\bibitem{attributesboosting}
T. Yao, Y. Pan, Y. Li, Z. Qiu, , and T. Mei.
\newblock Boosting image captioning with attributes.
\newblock In {\em ICCV}, 2017.

\bibitem{semanticatt}
Quanzeng You, Hailin Jin, Zhaowen Wang, Chen Fang, and Jiebo Luo.
\newblock Image captioning with semantic attention.
\newblock In {\em CVPR}, 2016.

\end{thebibliography}
}

\end{document}